\documentclass{article}

\PassOptionsToPackage{numbers, sort&compress}{natbib}
\usepackage[main, final]{neurips_2026}

\usepackage{amsmath,amsfonts,bm}

\def\eqref#1{equation~\ref{#1}}

\def\1{\bm{1}}

\DeclareMathAlphabet{\mathsfit}{\encodingdefault}{\sfdefault}{m}{sl}
\SetMathAlphabet{\mathsfit}{bold}{\encodingdefault}{\sfdefault}{bx}{n}

\usepackage[utf8]{inputenc}
\usepackage[T1]{fontenc}
\usepackage{url}
\usepackage{microtype}
\usepackage{graphicx}
\usepackage{subfigure}
\usepackage{booktabs}
\usepackage{amsmath}
\usepackage{amssymb}
\usepackage{mathtools}
\usepackage{amsthm}
\usepackage{float}
\usepackage{placeins}
\usepackage{tikz}
\usepackage{xcolor}
\usepackage{listings}
\usepackage{array}

\usetikzlibrary{shapes,arrows,positioning,fit,backgrounds,shadows,calc}
\usepackage[capitalize,noabbrev]{cleveref}
\usepackage[textsize=tiny]{todonotes}

\theoremstyle{plain}

\theoremstyle{definition}

\theoremstyle{remark}

\title{Belief Engine: Making Agent Stance Auditable in LLM Deliberation}

\author{\makebox[0.45\textwidth]{Joshua C. Yang\thanks{Work primarily performed while at ETH Zurich.}} \\
Massachusetts Institute of Technology \\
\texttt{joshyang@mit.edu}
\And
\makebox[0.45\textwidth]{Maurice Flechtner} \\
ETH Zurich and \\
Centre for Democracy Studies Aarau, \\
University of Zurich
\AND
\makebox[0.45\textwidth]{Damian Dailisan} \\
ETH Zurich
\And
\makebox[0.45\textwidth]{Michiel A. Bakker} \\
Massachusetts Institute of Technology
}

\begin{document}

\maketitle

\raggedbottom

\begin{abstract}
\hyphenpenalty=1000
\emergencystretch=1em
LLM-based AI agents are increasingly used to simulate deliberative social interactions, including negotiation, conflict resolution, and multi-turn opinion exchanges. Simulating human-like opinion change calls for agents that balance existing beliefs with openness to new evidence, while making their responses to that evidence traceable. Current systems often make this difficult: agents may drift from assigned roles, echo an interaction partner, keep a stance despite contrary evidence, or appear to change because the prompt or retrieved context has changed. We introduce the Belief Engine (BE), a framework that uses ``belief'' in a Bayesian modelling sense, as a maintained evidential state over a proposition that is exposed as a scalar stance. BE extracts structured evidence, stores argument records, and updates the belief state through a Bayesian-inspired stance update in log-odds space with two interpretable controls: evidence uptake $u$ and prior anchoring $a$. Across multiple LLM models, sweeps of $u$ and $a$ demonstrate stable control over stance dynamics. Higher uptake makes agents more responsive to new evidence, while stronger anchoring preserves the initial stance. On DEBATE, a human deliberation dataset with pre/post opinions, BE is most accurate when participants move in the direction of extracted evidence, showing its strength in evidence-following deliberative contexts. When all participants are evaluated together, gains are more modest because many people remain stable or move for reasons not captured by the extracted evidence stream. BE provides a configurable and auditable belief-update layer for evidence-grounded deliberation, in which LLM agents can be configured as open-minded, anchored, or evidence-sensitive, while their stance trajectories and memory can be audited, compared, and calibrated against human opinion change.

\end{abstract}

\begin{figure}[!t]
    \centering
    \includegraphics[width=\linewidth]{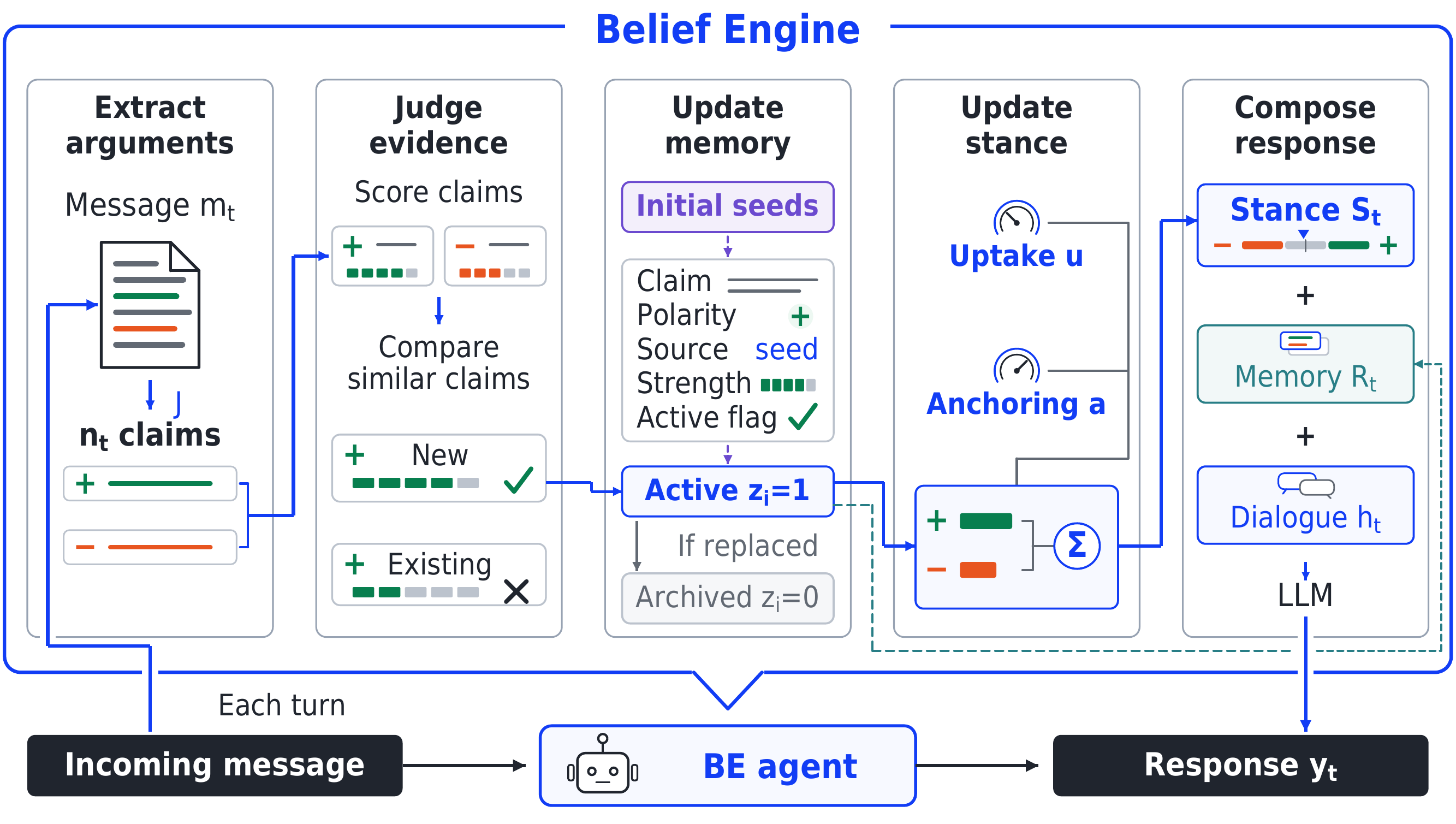}
    \caption{
        Belief Engine architecture. Incoming messages are converted into scored claims and consolidated in structured memory, retaining the stronger of similar claims with the same polarity. In the illustrated active-record mode, only active evidence contributes to stance: uptake $u$ weights later-message evidence, while anchoring $a$ weights initial seed evidence. Weighted contributions are summed in log-odds space and mapped to stance $S_t\in[-1,1]$, from opposition to support. The LLM uses this externally computed stance, retrieved memory, and recent dialogue to generate its reply. Solid blue arrows show per-turn processing; dashed purple and teal arrows show seed initialization and memory retrieval, respectively. Bars and the stance marker are schematic.
    }
    \label{fig:belief-engine}
\end{figure}

\section{Introduction}

Large language models (LLMs) are moving from single-turn assistants to simulated participants in agent societies, education, civic deliberation, and collective-decision settings. Generative-agent work showed how LLMs can sustain social simulation over time~\citep{park2023generative}, while recent deliberation and augmented-democracy studies use LLMs to model discussion, preference formation, and citizen input~\citep{Yang_2024,chuang2025debate,Gudino2024}. Digital-twin proposals extend this ambition by treating synthetic communities as test beds for deliberative design~\citep{novelli2025testingdeliberative}. Across these settings, researchers need to know what agents say and also whether, when, and why their stances change in response to reasons.

Because current LLM agents can generate fluent agreement and disagreement without exposing the source of that movement, attribution becomes central. A transcript may look more moderate because the agent accepted evidence, because its persona drifted~\citep{choi2025identitydrift}, because it echoed its interaction partner~\citep{shekkizhar2025echoing}, or because inherited model biases shaped the response~\citep{taubenfeldSystematicBiasesLLM2024}. An agent may also appear resistant because retrieval or prompt context changed. For agent-based simulations, it remains unclear what kind of deliberative behaviour is being represented \citep{flechtnerDeliberativeDeficitEmpirical2026}.

We introduce the Belief Engine (BE), an auditable simulation-control layer that puts log-odds belief updating under experimental control. We use ``belief'' in an operational Bayesian modelling sense, meaning a proposition-level evidential state maintained in log-odds. We use \textit{stance} for the scalar readout $S\in[-1,1]$, with positive values supporting the proposition and negative values opposing it. $S$ records only support for, or opposition to, one declared proposition. It does not capture partial agreement on sub-issues, trust in a speaker, or value trade-offs, so two agents can share the same $S$ for different reasons. Initial seed arguments define the prior, later arguments provide weighted evidence, and both are accumulated through a log-odds rule with two interpretable controls: \textit{evidence uptake}~$u$ and \textit{prior anchoring}~$a$. This targets the component of deliberation that current LLM agents leave implicit: what counts as evidence, how much it can move the belief state, and how strongly an initial commitment persists. Incoming arguments are extracted, judged, and stored in structured memory before they enter this update. Language generation is conditioned on the updated stance and retrieved evidence. 
Because $S$ is computed outside the generator and then supplied to it, the stance trajectory shows how BE updates and records its maintained state, not how an unconstrained LLM or a person would form beliefs. 
BE supports auditing by making the steps between a received message and a stance update available for inspection. Researchers can trace which arguments contributed to a change, examine how they were weighted, and check the resulting stance against the stated update rule.

Separating the maintained belief state from generation lets researchers state modelling assumptions that are usually buried in prompts. High uptake creates more evidence-responsive agents, strong anchoring preserves initial commitments, and the stance trajectory can be checked against the evidence that produced it. BE therefore turns deliberation simulation from prompt-driven transcript production into a reportable state-transition model in which researchers specify how evidence enters memory, how much it moves the belief state, and how the resulting stance conditions language.

We evaluate BE with controlled parameter sweeps, matched prompt-based baselines, and replay on 2{,}495 quality-filtered DEBATE~\citep{chuang2025debate} trajectories from human multi-round discussions. The sweeps test whether $u$ and $a$ control evidence responsiveness across base models. The prompt baselines test whether self-revision and retrieval expose the same update process. We use DEBATE replay to assess how closely the update rule reconstructs observed human stance changes from the extracted evidence participants received.

Concretely, BE makes three pieces of the simulation explicit. Its architecture separates how evidence is extracted, judged, and stored from how the stance is updated and expressed. In generated debates, two scalar controls, uptake and anchoring, produce predictable stance dynamics across base models while keeping an evidence-level audit trail. Replayed on human discussions, the same interface separates movement aligned with extracted evidence, stable anchoring, and movement that the extracted evidence does not account for.

\section{Related work}

\textbf{LLM deliberators blur expression and state.} A central difficulty in LLM-based social simulation is attribution. When an LLM agent changes what it says, the transcript alone does not show whether the movement should be treated as evidence-responsive updating or as a surface shift caused by prompting, retrieval, or generation \citep{flechtnerDeliberativeDeficitEmpirical2026}. LLM agents exhibit persona shift~\citep{choi2025identitydrift, shekkizhar2025echoing}, knowledge-conflict failures~\citep{xu2024knowledge}, recency sensitivity~\citep{kim2024bayesian, zhang2025discounted}, regression to training biases~\citep{taubenfeldSystematicBiasesLLM2024}, and excessive convergence relative to humans~\citep{chuang2025debate}. The DEBATE dataset is especially relevant because it records both public interaction traces and private pre/post opinions, showing that plausible role-play can still distort individual and group opinion dynamics. Debate can also behave like a martingale without a specified update policy~\citep{choi2025debateorvote}. Even if reasoning models internally simulate dialogic ``societies of thought''~\citep{kim2026reasoning}, those implicit perspectives are not persistent, auditable social actors. BE addresses this gap by maintaining a separate belief state with an explicit stance variable.

\textbf{Democratic simulations need inspectable change.} Applied systems increasingly use LLMs in deliberative settings. AI debate can support factual claim assessment~\citep{rahman2025aidebate} and judicial tasks~\citep{hu2025multiagent}. Agora and ArgueMate scaffold consensus-finding in civic and educational contexts~\citep{PradeepFulay2026, wang2026arguemate}, while Murmi supports participatory sensemaking in citizen deliberations~\citep{yang2026participatorysensemaking}. Augmented-democracy and digital-twin proposals extend this ambition by using LLMs or synthetic communities to estimate citizen preferences and test deliberative designs through controlled ``what-if'' scenarios~\citep{Gudino2024, novelli2025testingdeliberative}. DelibSim shows that LLM groups match human procedural discourse quality while failing to reproduce similar epistemic outcome dynamics~\citep{flechtner2026procedural}. We argue that, in such settings, the missing state variable becomes a substantive problem. A deliberative simulator should report final accuracy, epistemic and discourse quality, and user learning, but it should also expose why simulated participants move, remain stable, or polarise.

\textbf{Memory does not determine belief revision.} Agent memory systems decide what can be retained, reflected on, or retrieved. Generative Agents~\citep{park2023generative}, episodic memory banks~\citep{zhong2023memorybank}, reflection systems~\citep{xu2025amem, liu2024rmm}, graph memory~\citep{gutierrez2024hipporag, kang2025memoryos, huang2025licomemory}, large-scale simulations~\citep{piao2026agentsociety, xu2026topologyaware}, and retrieval-augmented debate~\citep{li2026rdebater} all strengthen some form of context persistence. But remembering an argument is not the same as accepting it as evidence. Memory can also amplify noise or reinforce experience-following behaviour~\citep{xiong2025memory}. BE places judgement and updating between memory and generation, so stored arguments become belief-relevant only when the evidence layer marks them active.

\textbf{Formal update rules need semantic grounding.} Bayesian models~\citep{olssonBayesianSimulationModel2013}, DeGroot updating~\citep{degroot1974reaching}, and bounded-confidence models~\citep{deffuant2000mixing, hegselmann2002opinion} make opinion dynamics explicit, and recent work connects LLM debate to Bayesian Nash equilibrium~\citep{xie2025econ}. Their abstraction is also their limitation for language-agent simulations. They rarely specify how natural-language arguments are extracted, judged, remembered, or turned back into utterances. Human belief updating also departs from ideal Bayesian assumptions~\citep{stengardGeneralityCognitiveBasis2022, holtUpdateBayesianUpdating2009, ashinoffEffectsBaseRate2022}. People may protect commitments, weigh evidence through identity and trust, and respond differently across social contexts. The BE framework sits between these traditions. It keeps the update rule parameterised, grounds each update in extracted arguments, and exposes profiles that can be calibrated against human deliberation trajectories.

\section{Method}
\label{sec:method}

Figure~\ref{fig:belief-engine} depicts a BE agent as an LLM-based debater whose proposition-level belief state is maintained by the Belief Engine. During a debate it exchanges messages with an opponent on a fixed topic. The agent can use any compatible base model. Across the generated-agent comparisons we use GPT-4o-mini, GPT-5.4-mini, Qwen~3.5 9B, and Gemma~4 E4B in different roles, as detailed below. In the generated-agent conditions, debates use a seed limit of $n{=}10$ arguments and use the same dialogue protocol
while varying the uptake--anchoring profile $(u,a)$. Each response is generated from three inputs: (i) current stance instruction, (ii) retrieved context, and (iii) recent dialogue context. Since the belief state is updated by reported parameters and exposed as $S$, the architecture makes it possible to study how simulated agents change state under argumentative pressure.

\subsection{Belief engine}
The Belief Engine decouples belief maintenance from generative reasoning so that the agents' belief updating can be controlled systematically. For each new message, the engine follows the five-step loop (Fig.~\ref{fig:belief-engine}): \textit{Extract Arguments} (convert a message into candidate claims), \textit{Judge Evidence} (score and merge near-duplicate candidates), \textit{Update Structured Memory} (store active and archived evidence), \textit{Update Belief State} (apply the log-odds rule and compute stance $S$), and \textit{Compose Response} (retrieve active memory in proportion to its pro/con composition and map $S$ to behavioural instructions for generation). Appendix~\ref{sec:component-comparisons} reports targeted checks of component behaviour and their scope.

As shown in Fig.~\ref{fig:belief-engine}, the Belief Engine uses a deterministic log-odds updater that recomputes the belief state from argument polarity, support strength, evidence uptake $u$, and prior anchoring $a$, then exposes it as stance $S$. Each stored argument has an \textit{active} flag $z_i \in \{0,1\}$. Only active records contribute to stance $S \in [-1,1]$, separating record-level evidence selection from the aggregate stance.

\paragraph{Extract arguments.}\label{sec:argument-extraction}
An independent LLM module decomposes each message $m_t$ into candidate records $\tilde{\mathcal{E}}_t=J(m_t)=\{\tilde e_i=(c_i,p_i,r_i)\}_{i=1}^{n_t}$, where $c_i$ is the canonical claim, $p_i\in\{-1,+1\}$ is its polarity ($+1$ affirmative, $-1$ negative), and $r_i\in\{\texttt{seed},\texttt{self},\texttt{opponent}\}$ is the source role. We use the same extractor across experiments, while the base debater model varies in generated-agent comparisons.
Polarity is defined relative to the debate proposition, independent of sentiment. %
A single message may yield zero, one, or several candidate arguments. Extraction only proposes candidate evidence records, and the judgement step decides whether a claim affects the belief state. We adopt a broad conception of deliberative communication that includes narratives, rhetorical forms, and experiential knowledge~\citep{youngInclusionDemocracy2002, nakazawaEffectStorytellingAttitude2024}. Accordingly, any persuasive contribution, whether formal reasoning, anecdotal evidence, or rhetorical framing, is treated as a valid unit of analysis. The extractor prompt and its output schema are listed in Appendix~\ref{sec:exact-prompts}. Appendix~\ref{sec:judgement-validation-details} reports validation details for the extractor and judgement layers.

\paragraph{Judge evidence.}\label{sec:judgement-layer}
The judgement layer governs how candidate arguments are transformed into structured evidence. Following \citet{xiong2025memory}, we treat scoring and conflict resolution as separate architectural choices with explicit outputs. \textbf{Strength scoring.}
Each extracted argument receives a strength $s_i\in[0,1]$ from either the LLM extractor ($s_i^{\mathrm{LLM}}$) or a classifier ($s_i^{\mathrm{clf}}$). The classifier is \texttt{DeBERTa-v3-large}~\citep{deberta2023paper} fine-tuned as a regressor on crowd-rated argument-quality labels~\citep{Gretz2019argumentquality}, mapping each claim text to a scalar score in $[0,1]$. Appendix~\ref{sec:code-release-plan} gives the released checkpoint and scoring procedure. We set $s_i=s_i^{\mathrm{clf}}$ when classifier scoring is enabled and $s_i=s_i^{\mathrm{LLM}}$ otherwise. Sweeps may use either scorer, while DEBATE replay fixes $s_i=s_i^{\mathrm{clf}}$ so calibration varies only belief-update parameters. Each record is then written as $e_i=(c_i,p_i,s_i,r_i,z_i)$ with $z_i\leftarrow 1$ prior to conflict resolution. \textbf{Conflict resolution.}
To avoid similar arguments causing repeated updates, we soft-deduplicate same-polarity active claims. For a new argument, we compare its claim embedding against active records with the same polarity. If none exist, or if the nearest match has cosine similarity below the configured threshold $\theta$, the new record stays active. Otherwise, we keep only the stronger record active and archive the other. Records are retained for auditability, but in this mode only active ones ($z_i=1$) affect the belief state. Appendix~\ref{sec:deduplication-details} gives the replacement rule and reports how $\theta$ is set.

\paragraph{Update structured memory.}\label{sec:memory-layer}
After conflict resolution, the BE agent stores each claim as a structured record with polarity, strength estimates, source role, and an active flag. Accepted evidence remains active, while superseded evidence is archived for traceability. This selective consolidation supports record-level replacement during conflict resolution and later retrieval by active pro/con memory composition.

\paragraph{Update belief state.}\label{sec:belief-updating}
The Belief Engine exposes an uptake--anchoring (UA) profile $(u,a)$. \textit{Evidence uptake} determines how strongly new evidence shifts the belief state, while \textit{prior anchoring} sets the strength of the initial prior.

Belief updating is modelled as additive evidence accumulation in log-odds space over the currently active memory, with bounded stance $S_t$. Let $\mathcal{A}_t=\{i:z_i=1\}$ denote the active-memory index set after judgement and conflict resolution at time $t$. The log-odds state is recomputed from the active set, so arguments that have since been archived no longer contribute. With $\gamma_i=a$ for seed records ($r_i=\texttt{seed}$) and $\gamma_i=u$ for later debate records ($r_i\neq\texttt{seed}$), the update is
\begin{equation}
    L_t = \sum_{i\in\mathcal{A}_t}
    p_i \ln\!\left(1+s_i\,\gamma_i\right),
    \qquad
    S_t = 2\sigma(L_t)-1 = \frac{2}{1+\exp(-L_t)}-1,
    \quad S_t\in[-1,1],
    \label{eq:belief-update}
\end{equation}
where $\sigma(x)$ is the logistic function. 
Seed records represent the prior, with no separate intercept. The anchoring $a$ sets how strongly seed records establish the initial belief state, while $u$ sets how strongly later debate evidence shifts it. Evidence accumulates linearly in log-odds, while the logistic transform yields bounded stance and diminishing returns near certainty. This update draws on the Bayesian idea of combining prior beliefs with new evidence in log-odds space. Argument scores determine how strongly each argument contributes to the updated belief, without requiring a calibrated likelihood model. In this mode, archived records remain auditable but no longer affect the belief state. 
Because $a$ scales the seed terms once while $u$ scales every later evidence increment, different $(u,a)$ profiles can reach the same final stance along different intermediate trajectories. BE therefore records the full stance path so that such differences remain visible. Appendix~\ref{sec:endpoint-paths} reports a fixed-evidence analysis of matching final stances and different intermediate trajectories.

\paragraph{Compose response.}\label{sec:retrieval-conditioning}
After updating the belief state, the BE agent retrieves a bounded context $R_t$ from active memory. Let $\mathcal{M}_t^+$ and $\mathcal{M}_t^-$ denote the active affirmative and negative records. Retrieval allocates $k_+=\operatorname{round}\!\left(k|\mathcal{M}_t^+|/(|\mathcal{M}_t^+|+|\mathcal{M}_t^-|)\right)$ slots to affirmative evidence and $k_-=k-k_+$ to negative evidence, with an even split when active memory is empty. $R_t$ is the union of the strongest $k_+$ and $k_-$ active records by support strength. Retrieval therefore follows accepted evidence and does not directly amplify the scalar stance. The next utterance is generated as $y_t \sim \mathrm{LLM}(S_t,R_t,h_t)$, where $S_t$ is mapped to a 10-bin stance-intensity instruction and $h_t$ is recent dialogue context. 
The generator receives $S_t$ as an input and does not directly modify it, so any later change in stance must pass through the configured update rule.

\subsection{Experimental setup}
\label{sec:experimental-setup}

We separate two empirical targets. Generated-agent experiments evaluate the full loop, including generation, judgement, memory, retrieval, and belief updating. DEBATE replay isolates the update rule under real received-evidence histories, giving a human-grounded calibration test for the resulting stance dynamics.

\paragraph{Generated-agent experiments.}
Seed and counter-agent arguments come from the Argument Quality Ranking dataset \citep{Gretz2019argumentquality}, which contains about 30{,}000 crowd-annotated arguments with topic, polarity, and quality scores. All generated debates run for 15 rounds with a limit of $n{=}10$ seed arguments per side and agent temperature $T{=}0.7$. Symmetric two-agent debate provides a controlled setting for testing update mechanics, profile dynamics, and evidence-conditioned generation.

In Secs.~\ref{sec:sensitivity} and \ref{sec:prompt-based-baselines}, we focus on (i) five-point parameter sweeps over $u$ or $a$ across three base models (five trials per setting) and (ii) matched prompt baselines, namely prompt-only and RAG. Appendix diagnostics add Open-minded/Stubborn profile and topic-grid demonstrations over 10 topics with three debates per profile-pairing cell. When a system does not expose $S$, a shared external LLM judge receives the proposition and generated text, then returns a scalar stance in $[-1,1]$ at temperature $0.0$. Appendix~\ref{sec:repro-prompts-models} collects the implementation details, with settings for the earlier diagnostics in Table~\ref{tab:hyperparams} and model roles in Table~\ref{tab:model-identifiers}. The common-evaluator comparison uses the separate protocol in Appendix~\ref{sec:common-evaluator-details}.

\paragraph{Human replay validation.}
We use the DEBATE dataset as a baseline for observed-evidence replay \citep{chuang2025debate}. This differs from its original simulation tasks. For each participant, we initialise BE from the private initial Likert stance, replay directed evidence from partner tweets and received chat messages, and compare the final BE stance with the private final Likert stance. Six-point Likert responses use fixed log-odds anchors $(-3,-1.5,-0.5,0.5,1.5,3)$, mapped to stance by $S=2\sigma(L)-1$. In this replay, the participant's initial stance supplies the prior in place of seed arguments. Anchoring $a$ scales the log odds of that initial stance, while uptake $u$ controls the contribution of received evidence. We perform a grid search over $(u,a)$ and select settings by training-fold RMSE, then evaluate final-stance RMSE on the held-out fold. We report two references. A no-change baseline predicts the final stance from the initial stance, and a net-evidence linear baseline, $\hat S_{\mathrm{final}} = S_{\mathrm{initial}} + \beta E$, fits one scalar evidence weight on training folds only. The linear baseline uses the same quality-filtered, received-evidence stream, human-like uptake and deduplication filter, and classifier strength scores as BE, but omits the log-odds belief-update rule. The replay excludes self-authored posts and messages, isolating the belief-update component under the evidence a participant received.

\section{Results}

\noindent The results ask three questions. First, is the maintained stance $S$ visible in generated text strongly enough to support comparison with external stance scores? Second, do uptake and anchoring give predictable control across base models, and do Prompt-only or RAG memory expose a comparable update trail? Third, when replayed on DEBATE, which uptake--anchoring profiles best match observed final stances under the received-evidence stream? Generated-agent experiments test controllability and auditability, while DEBATE replay tests how the same update interface captures human response heterogeneity. 

\subsection{Uptake and anchoring control stance dynamics}
\label{sec:sensitivity}

We first validate that the maintained stance $S$ is expressed in generated language. We sweep $S \in [-1,1]$ across ten propositions and ask an external judge to score each generated response. The scores align tightly with the assigned stance ($r = 0.959$, topic-clustered 95\% CI $[0.952,0.964]$), with a fitted slope of $0.928$ indicating mild compression at the extremes. Because the judge sees the proposition and response but not $S$, this checks how closely the generated text expresses the assigned stance. 
This shows that the externally computed stance shapes observable language. Whether human readers perceive the same stance remains to be tested. Appendix~\ref{sec:stance-alignment-details} reports the protocol, per-topic results, and robustness checks.

\begin{figure}[H]
    \centering
    \begin{minipage}[t]{0.61\linewidth}
        \vspace{0pt}
        \centering
        \includegraphics[width=\linewidth]{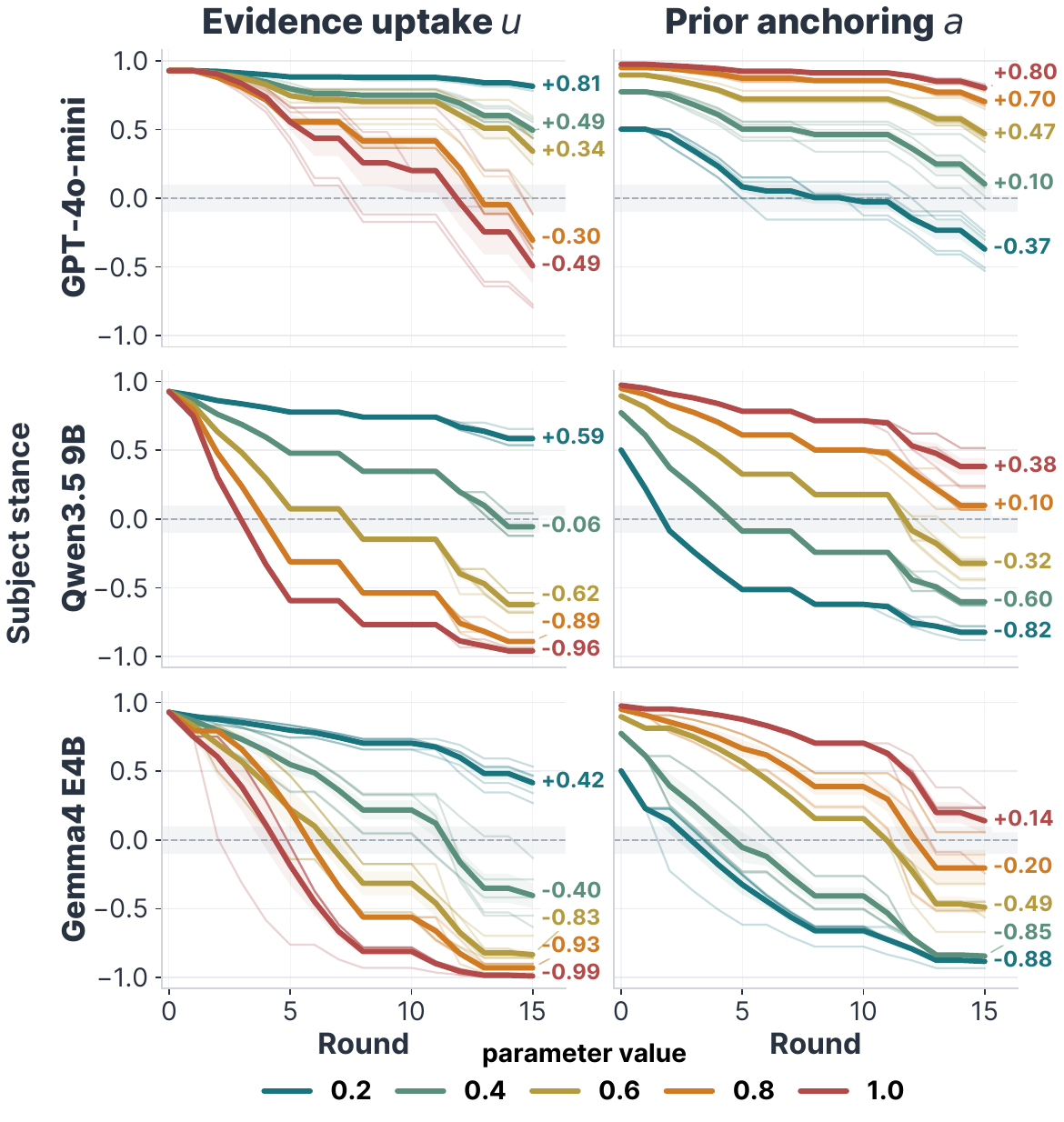}
    \end{minipage}
    \hfill
    \begin{minipage}[t]{0.35\linewidth}
        \vspace{0pt}
        \centering
        \includegraphics[width=\linewidth]{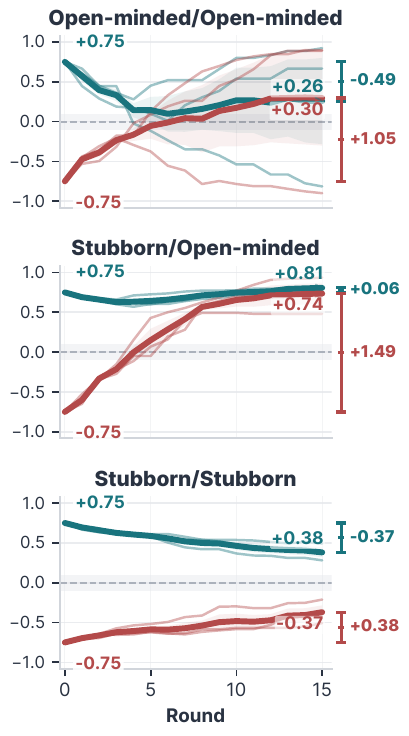}
    \end{minipage}
    \small
    \caption{Parameter control and profile dynamics. Stance $S\in[-1,1]$, where $+1$ means affirmative with respect to the proposition and $-1$ means negative. \textbf{Left:} BE single-agent sweeps across three base LLMs, varying evidence uptake $u$ ($a=0.70$) and prior anchoring $a$ ($u=0.4$). \textbf{Right:} GPT-5.4-mini two-agent profile debates on austerity with Open-minded $(u,a)=(0.40,0.20)$ and Stubborn $(0.10,0.80)$ agents. Thin lines show trials, thick lines show means.}
    \label{fig:model-sensitivity-be-agent-trajectories}
\end{figure}

We next test the two controls directly. These single-agent sweeps admit eight seed arguments and use the historical incremental update mode (Appendix~\ref{sec:update-semantics}). We sweep \textit{evidence uptake} $u$ and \textit{prior anchoring} $a$ across five levels ($\{0.2, \dots, 1.0\}$) in 15-round debates against a deterministic counter-argument opponent on the proposition ``\textit{We should introduce compulsory voting}''. We repeat the sweep with three base language models (GPT-4o-mini, Qwen~3.5 9B, and Gemma~4 E4B) to test whether the pattern depends on the generator. When we vary evidence uptake, all agents start at the same stance, approximately $+0.93$. When we vary anchoring, stronger anchoring gives more weight to the initial seed arguments, so agents begin with stronger support for the proposition. Figure~\ref{fig:model-sensitivity-be-agent-trajectories} shows the intended ordering.

Higher $u$ makes agents more responsive to new evidence. For GPT-4o-mini, the final stance shifts from $+0.81$ ($u{=}0.2$) to $-0.49$ ($u{=}1.0$), and Qwen and Gemma move even further in the high-uptake setting ($-0.96$ and $-0.99$). Higher $a$ makes the seeded prior more persistent. Holding $u{=}0.4$, GPT-4o-mini ranges from $-0.37$ ($a{=}0.2$) to $+0.80$ ($a{=}1.0$), and Qwen and Gemma follow the same pattern.

Figure~\ref{fig:model-sensitivity-be-agent-trajectories} (right) shows that the same rule could also create recognisable two-agent profiles. Open/Open agents reduce disagreement, Stubborn/Open pulls the open con agent toward the anchored pro agent, and Stubborn/Stubborn preserves more of the initial gap. The outcomes differ by base model and trial because retrieval keeps active evidence from both sides available, so generation still affects the rate and magnitude of movement. This is the desired behaviour for simulation, because openness and anchoring become explicit settings that researchers can report.

\FloatBarrier
\subsection{Prompt-based self-updating produces less convergence than BE}
\label{sec:prompt-based-baselines}

We compare BE with two prompt-based baselines: \textit{Prompt self-update}, which asks agents to reconsider their stance after each opponent message, and \textit{RAG+self-update}, which additionally supplies retrieved arguments when generating replies. We run ten trials per method on each of five topics, for a total of 150 debates. In each debate, two agents start at opposing stances of ${\pm}0.75$ and exchange messages for 15 rounds. All methods use the same base model and response budget. A shared external judge scores their responses without seeing the method labels.

By the final round, BE agents express positions that are closer together than those of either baseline. The average final stance gap is $0.53$ for BE, compared with $1.12$ for Prompt self-update and $1.02$ for RAG+self-update. A gap of $0$ means that the judge assigns both responses the same stance score. A gap of $2$ places them at opposite extremes. BE has smaller gaps on all five topics, and the aggregate pattern persists when each topic is excluded in turn. Confidence intervals and per-topic results appear in Appendix~\ref{sec:common-evaluator-details}. These results show that BE agents reach closer expressed positions under the tested protocol.

\begin{figure}[!htbp]
    \centering
    \begin{minipage}[t]{0.43\linewidth}
        \centering
        \includegraphics[width=\linewidth]{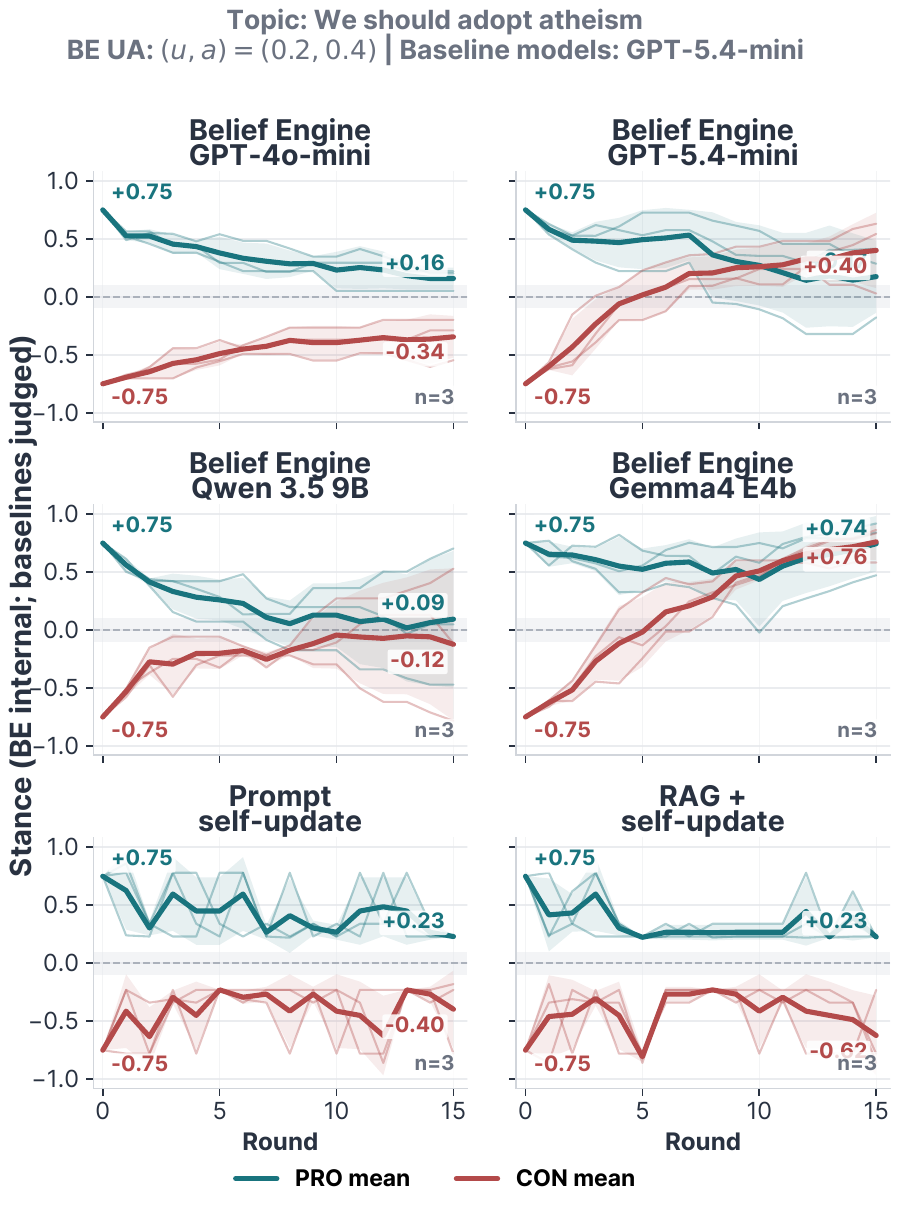}
    \end{minipage}
    \hfill
    \begin{minipage}[t]{0.56\linewidth}
        \centering
        \includegraphics[width=\linewidth]{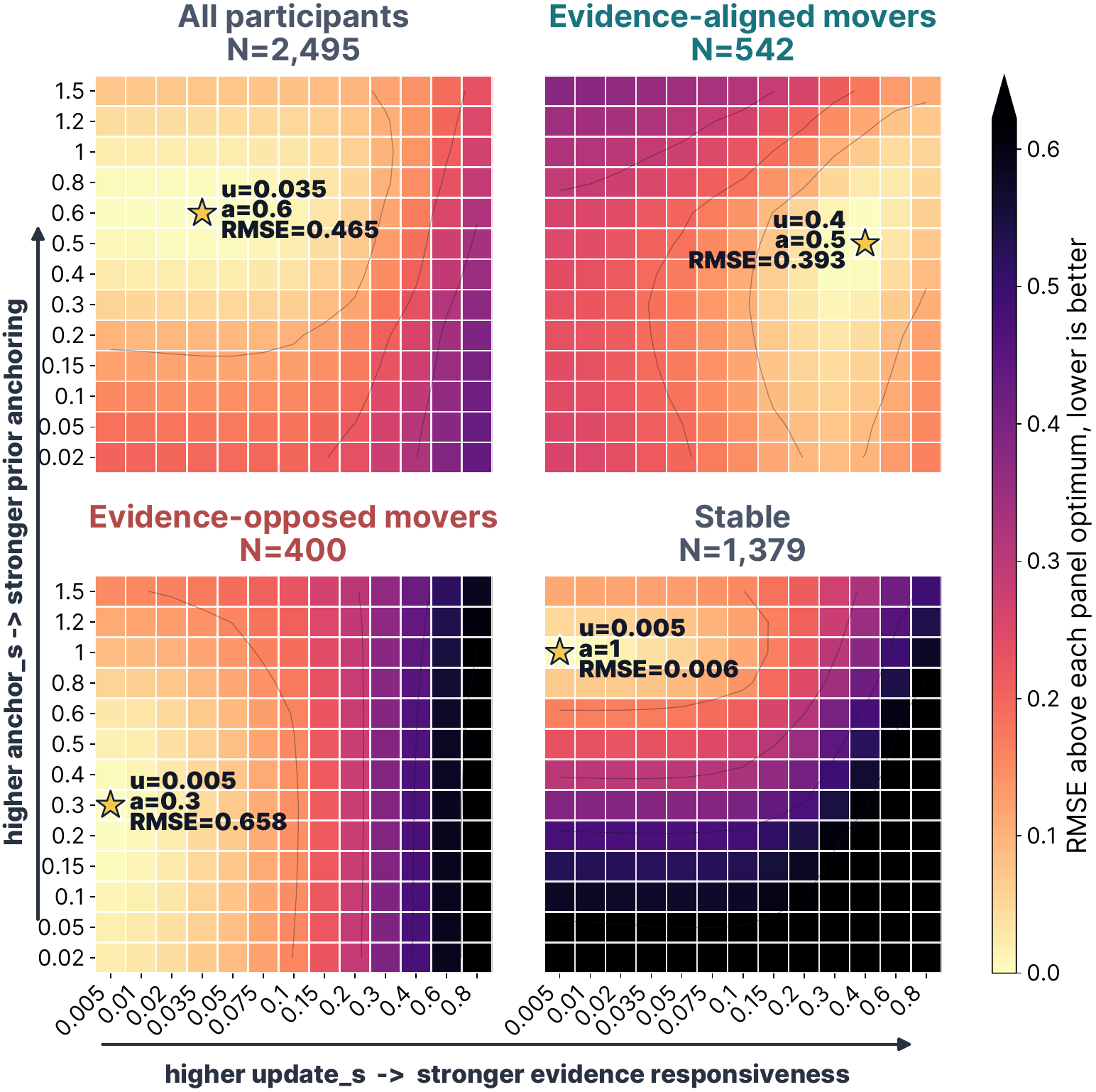}
    \end{minipage}
    \caption{Mechanism and calibration checks.  \textbf{Left:} Earlier diagnostic: 15-round atheism debates compare BE variants (fixed $(u,a)=(0.2,0.4)$) with GPT-5.4-mini and RAG baselines. BE panels show internal stance and baselines show external-judge scores. These mixed measurements are not the primary cross-method comparison. \textbf{Right:} DEBATE replay sweeps evidence uptake $u$ (x-axis) and prior anchoring $a$ (y-axis). Each heatmap is one participant subset. Cells show mean training RMSE across five group-split and five topic-split training folds, relative to the panel minimum. Lower values are better. A star marks the panel optimum.}
    \label{fig:mechanism-and-calibration}
\end{figure}

\subsection{Human replay reveals different uptake--anchoring profiles}
\label{sec:evidence-alignment-parameters}

The DEBATE replay keeps the received evidence fixed and tests which uptake--anchoring settings best reconstruct participants' final stances. We compare groups defined by how their opinions actually changed. These groups help us interpret the observed outcomes, but cannot be used to assign profiles before the final opinions are known. Figure~\ref{fig:mechanism-and-calibration} (right) shows the calibration surface. Each heatmap sweeps $u$ and $a$ for one participant subset and colours each cell by how much its mean training RMSE exceeds that panel's best cell. The pooled surface favours low uptake with moderate anchoring. After partitioning by evidence alignment, the minima move to different regions: evidence-aligned movers favour high uptake, evidence-opposed movers favour near-zero uptake, and stable participants favour near-zero uptake with maximal anchoring. The panel therefore shows why a single default profile is misspecified for a mixed population.

\begin{table}[H]
\centering
\caption{Pooled held-out RMSE in reconstructing final human stances on the $[-1,1]$ scale. Lower RMSE is better. \textit{No-change} carries the initial stance forward. \textit{Net-evidence} uses the same signed extracted evidence as BE, with a scalar weight fitted on the training folds. \textit{BE} uses each group's calibrated uptake--anchoring settings. Mean $|\Delta|$ is the average absolute pre/post human stance change.}
\label{tab:debate-main-rmse}
\setlength{\tabcolsep}{3pt}
\begin{tabular}{lrrrrr}
\toprule
\textbf{Group} & \textbf{$N$} & \textbf{Mean $|\Delta|$} & \textbf{No-change} & \textbf{Net-evidence} & \textbf{BE} \\
\midrule
All participants & 2{,}495 & 0.283 & 0.489 & 0.488 & 0.466 \\
Evidence-aligned movers & 542 & 0.619 & 0.718 & 0.689 & 0.395 \\
Evidence-opposed movers & 400 & 0.633 & 0.737 & 0.765 & 0.662 \\
Stable participants & 1{,}379 & 0.000 & 0.000 & 0.037 & 0.006 \\
\bottomrule
\end{tabular}%
\end{table}

Table~\ref{tab:debate-main-rmse} reports the held-out errors and mean observed movement. In the pooled population, BE reduces RMSE from $0.489$ under no-change and $0.488$ under net-evidence linear replay to $0.466$. Outcome-conditioned diagnostic gains are larger. Evidence-aligned and evidence-opposed movers have similar movement magnitudes (mean $|\Delta|=0.619$ and $0.633$), but require different profiles. Aligned movers are reconstructed by high uptake ($0.395$ RMSE), while opposed movers flag cases where extracted evidence points against observed movement. Stable participants do not show predictive improvement over no-change, which is exact by definition for this group. Instead, they check that BE can represent resistance to extracted evidence without forcing movement. The group-specific settings have lower RMSE than the prior-only and pooled global replay references in each of the five folds, both when holding out discussion groups and when holding out topics (Appendix~\ref{sec:debate-fold-robustness}). This supports using different profiles to reconstruct the observed response patterns.

\begin{figure}[!htbp]
    \centering
    \includegraphics[width=\linewidth]{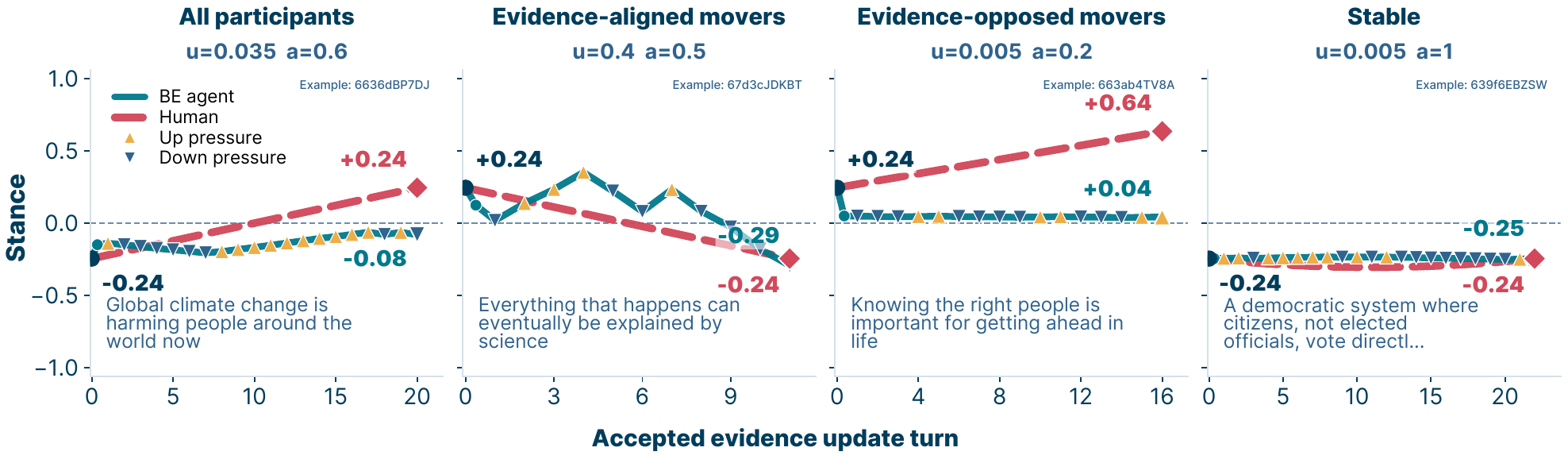}
    \caption{
        Single-participant DEBATE replay examples. Each panel samples one participant from a diagnostic subset and applies that subset's calibrated $(u,a)$ profile, shown above the panel. The x-axis counts accepted received-evidence updates rather than debate rounds. Starting from the observed human initial stance, the teal curve shows the BE replay after each accepted evidence item. Markers indicate whether the extracted item pushes stance upward or downward. The red dashed line connects only the observed human initial and final stances, since DEBATE does not measure intermediate human stances.
    }
    \label{fig:debate-individual-trajectories}
\end{figure}

At the single-participant level, Fig.~\ref{fig:debate-individual-trajectories} provides explanatory traces for the update mechanism. Aligned endpoints show cases where accepted evidence accounts for movement, while divergence points to interpretation, social context, or arguments not captured by extraction. These patterns separate three cases that would look similar in a Prompt-only agent: movement with accepted evidence, stability under evidence pressure, and movement that the extracted evidence cannot explain.

\FloatBarrier
\section{Discussion and future work}

These experiments suggest that the value of BE is not to make every deliberative trajectory predictable, but to make change attributable. ``Belief'' here is an operational evidential state over a proposition, maintained in log-odds and exposed through scalar stance. Under this definition, researchers can specify what counts as evidence, how strongly it shifts the belief state, and how strongly initial commitments persist. The judgement layer validation shows reliable directed-evidence extraction on clean single-argument inputs (96.9\% polarity accuracy), DEBATE replay improves over no-change and net-evidence linear references while showing why a single profile is insufficient for a heterogeneous population, and the prompt comparison shows that externally scored stance change does not expose the update process itself.

The same operational definition limits what BE can claim to model. Because the stance is computed by a researcher-chosen rule and imposed on the generator, observed stance changes largely reflect that rule by design. Our claims therefore concern the control and auditing of stance change and leave intrinsic LLM belief updating as a separate question. A single proposition-level stance also cannot capture every way people deliberate. They may partly agree, trade off values, distrust a speaker, or soften rhetorically without moving on the headline proposition.

Some of this granularity is already present. Each argument record is tied to its own claim, so a new argument on an earlier claim can replace that record and shift the stance through it. Trust is partly reflected in uptake $u$, although isolating it as its own parameter would need a separate mechanism. A natural extension is a multidimensional state in which each argument updates the sub-issue it affects, and retrieval brings back the records relevant to that sub-issue. How these dimensions should combine into the headline position, and which datasets could calibrate them, remain open questions.

The DEBATE replay further clarifies what calibration means for deliberative agents. We do not expect one global profile to match everyone, and averaging all participants together can hide behaviours that a heterogeneous simulator may need to represent. Some people stay close to their initial stance, some move with the extracted evidence they receive, and others move in ways the current evidence stream does not explain. BE makes these differences explicit: a stable endpoint can be represented as anchoring, evidence-aligned movement as uptake, and divergence from extracted evidence as a signal that some social or interpretive factor is missing. Which profile is appropriate depends on the deliberative setting. Our point is that BE lets these profiles be separated, inspected, and reported.

DEBATE captures one particular deliberative context. Its participants encounter partner posts and chat messages in a short online setting, and most do not show mapped Likert movement. That stability is informative for online opinion-exposure simulations, but it should not be confused with settings such as citizens' assemblies, facilitated workshops, or in-person deliberation, where participants may enter with an explicit norm of listening, compromise, and joint problem solving. Separating uptake, anchoring, and response profiles lets BE change the assumed deliberative context. A citizens' assembly simulation, for example, may contain more evidence-responsive or compromise-seeking agents, or agents whose movement appears first on sub-issues and value trade-offs rather than on the headline proposition.

All our experiments are two-agent discussions. Although DEBATE participants interacted in four-person groups over pairwise rounds, we treat each participant's data as a two-agent discussion, without distinguishing between speakers. Since the update rule does not depend on how many people an agent talks to, future work can extend BE to generated multi-party or longer deliberations, though this needs further design choices such as when an agent should interject.

Finally, evidence assessment remains a modelling choice. BE makes this choice explicit, but it does not make it value-free. LLM extractors, base models, external judges, and the AQR-trained strength classifier can all reflect training-data, safety-tuning, and community-specific priors. This matters most for moral and political topics, where argument quality is contested. For new domains, the judgement layer should be calibrated against domain data and, where appropriate, audited with human or participatory feedback. BE makes these choices available for auditing. When a trajectory is unexpected, researchers can examine which claims were extracted, which records remained active, and how their weights shaped the update. This helps locate whether the outcome arose from the interpretation of evidence, its selection in memory, or the configured update rule, providing a basis for targeted changes to the simulation.

The broader implication is that a simulated persona should specify how the agent treats new information, in addition to its role. A prompt can say who an agent is, while a memory-and-judgement process specifies what the agent treats as experience, which parts of that experience count as evidence, and how much they can revise persistent commitments. BE moves deliberative-agent simulation toward this standard, with accountable trajectories in which convergence, polarisation, and stability can be traced back to explicit assumptions.

\section{Conclusion}

As LLM agents move from single-turn assistants into deliberative multi-agent systems, stance change becomes a modelling problem. We presented the Belief Engine, a framework that separates evidence extraction, judgement, memory, log-odds updating, stance computation, and response generation. By combining persistent memory with a belief state that remains open to new evidence, BE offers a configurable basis for modelling the human-like balance between retaining prior beliefs and revising them. Researchers can configure evidence uptake and prior anchoring, then audit stance changes by tracing which arguments contributed, how they were weighted, and whether updates followed the specified rules. Across tested base models, these controls make agents predictably more evidence-responsive or more anchored. DEBATE replay shows that BE best reconstructs human opinion changes aligned with the extracted evidence, supporting its use for simulating evidence-responsive behaviour. BE therefore provides a foundation for configurable agent simulations in which researchers can audit how persistent beliefs change in response to new evidence.

\newpage

\begin{ack}
We gratefully acknowledge the Computational Social Science team at ETH Zurich, where Joshua C. Yang, Maurice Flechtner, and Damian Dailisan were employed while conducting this work, and the Horizon Europe project AI for Deliberation, which partially supported Maurice Flechtner. The authors declare no competing interests.
\end{ack}

\clearpage
\appendix

\setcounter{figure}{0}
\renewcommand{\thefigure}{A\arabic{figure}}
\setcounter{table}{0}
\renewcommand{\thetable}{A\arabic{table}}

\raggedbottom
\section{Design Rationale and Component Comparisons}
\label{sec:design-rationale-details}

\begin{table}[H]
\centering
\caption{Design rationale for the Belief Engine. Each component gives researchers a separate handle on belief-state change in deliberative-agent simulations.}
\label{tab:design-rationale}
\small
\setlength{\tabcolsep}{5pt}
\renewcommand{\arraystretch}{1.12}
\begin{tabular}{@{}>{\raggedright\arraybackslash\bfseries}p{0.19\linewidth}
>{\raggedright\arraybackslash}p{0.28\linewidth}
>{\raggedright\arraybackslash}p{0.18\linewidth}
>{\raggedright\arraybackslash}p{\dimexpr0.35\linewidth-30pt\relax}@{}}
\toprule
\textbf{Design choice} & \textbf{Rationale} & \textbf{Alternative} & \textbf{Risk addressed} \\
\midrule
External belief state & Keeps belief maintenance separate from response generation. &
Prompt-only persona or hidden context. & Reduces untraceable drift and context-dependent stance changes. \\
\addlinespace[5pt]
Argument-level evidence & Stores claims, polarity, strength, and provenance as structured records. &
Free-form memory summaries. & Preserves auditability and enables record-level ablations. \\
\addlinespace[5pt]
Judgement layer & Makes extraction, polarity, and strength estimation explicit modules. &
Let the generator implicitly decide what mattered. & Exposes where bias, calibration error, or disagreement enters. \\
\addlinespace[5pt]
Active flag & Marks records eligible for stance updates and retrieval. &
Delete weak or conflicting memories. & Preserves rejected/superseded records for inspection. \\
\addlinespace[5pt]
Log-odds update rule & Provides ordered, bounded, reproducible belief-state updates with interpretable parameters and scalar stance. &
Direct LLM numerical updates. & Avoids stochastic jumps and makes uptake/anchoring tunable. \\
\addlinespace[5pt]
Memory-composition-proportional retrieval & Conditions future responses on accepted evidence while keeping minority-side active memories eligible for retrieval. &
Stance-conditioned or semantic-only retrieval. & Avoids making the current stance itself determine evidence exposure, while still linking accepted memory to observable language. \\
\bottomrule
\end{tabular}
\end{table}

\subsection{Evidence on component behaviour}
\label{sec:component-comparisons}

Table~\ref{tab:design-rationale} explains the intended function of each component. The comparisons below test particular consequences of those choices. Final disagreement, agreement with human quality ratings, and resistance to repeated evidence assess different properties, so no single outcome validates the whole design.

\paragraph{Argument decomposition.}
In a focused comparison, storing separate reasons rather than one whole-message record reduced final externally judged disagreement by $1.086$ (adjusted interval $[0.995,1.171]$). The comparison covers ten topics and 50 matched topic--seed blocks. Its direction is consistent across every topic and when each topic is excluded in turn. Both conditions retain memory, but structured extraction produces an average of 6.34 claims per message, compared with exactly one proposition-classified verbatim record. The result shows that this decomposition procedure changes expressed stance dynamics. It does not isolate the structure of the records from the number of evidence contributions or directly measure auditability.

\paragraph{Other effects on disagreement.}
Agents that kept evidence from earlier turns ended with a smaller stance gap than agents using only the latest opponent message. The estimated reduction was $0.165$, with an adjusted interval of $[0.052,0.286]$. For the other components, we also measure how much enabling each one reduces the final gap. The estimates were $-0.024$ for response retrieval ($[-0.114,0.062]$), $+0.014$ for deduplication ($[-0.355,0.370]$), and $-0.048$ for graded rather than fixed strength scores ($[-0.367,0.240]$). Each of these three intervals includes zero, so the study does not clearly establish whether these choices increase or decrease final disagreement.

An uncertain effect on disagreement does not show that a component makes no difference or can be removed. Components may serve other purposes. For example, this study does not measure whether retrieval helps responses use relevant evidence. It also does not establish that closer final positions mean better deliberation.

\paragraph{Checks tied to component functions.}
For deduplication, a controlled repeated-claim test directly checks the intended protection against repeated counting: stance stayed unchanged when equivalent evidence was repeated three or five times with deduplication enabled. This is a controlled invariance result, not a measure of general duplicate-detection accuracy. In a separate natural-dialogue replay, removing deduplication modestly improved squared error. For scoring, Appendix~\ref{sec:judgement-validation-details} reports agreement with crowd-rated argument quality, rather than an effect on convergence. Earlier replay comparisons held accepted evidence fixed and varied scoring and the update rule. Uniform versus classifier strengths gave inconclusive RMSE differences. Linear stance aggregation also gave inconclusive RMSE differences relative to log-odds updating, while reducing MAE by $0.0198$. These checks do not establish a uniquely superior scorer or update rule, or equivalence in generated dialogue.

\paragraph{Focused-study protocol.}
The study crosses separate versus whole-message records, retrieval on versus off, and persistent versus turn-local memory, giving eight conditions. Two further conditions disable deduplication or replace graded strengths with the fixed value $0.8476$ at the full setting. Five trials per condition on ten topics give 500 debates. All use GPT-5.4-mini, 15 rounds, initial stances $\pm0.75$, uptake $u=0.2$, and the explicit-prior mode in Appendix~\ref{sec:update-semantics}, where anchoring has no numerical effect. Responses use four recent messages and up to ten retrieved records when retrieval is enabled. Turn-local memory retains only evidence from the latest opponent turn alongside the fixed prior. Uniform scoring changes new-record strengths and preserves seed strengths. The Pro agent speaks first, and a shared external GPT-5.4-mini evaluator scores the responses.

Decomposition, retrieval, and persistence contrasts average over the other crossed choices. Deduplication and scoring are tested only at the full setting, without identifying their interactions. All five intervals above use 100,000 hierarchical bootstrap draws over topics and then matched trial blocks (analysis seed 20260725). The 99\% percentile intervals give nominal 95\% familywise coverage using a Bonferroni adjustment. These generated-debate results remain separate from earlier fixed-stream replay and three-condition persistence/retrieval checks. Their scope is expressed disagreement under one model family and a fixed speaker order, not truthfulness or human realism.

\flushbottom
\section{Conflict-Resolution Details}
\label{sec:deduplication-details}

The main text describes conflict resolution at a high level. Here we give the exact rule used to decide whether a candidate argument remains active. Let $\phi(c)$ be the embedding of claim $c$ and define cosine similarity as
\[
    \operatorname{sim}(i,j)
    =
    \frac{\phi(c_i)^\top \phi(c_j)}
    {\|\phi(c_i)\|\,\|\phi(c_j)\|}.
\]
For a new argument $e_i$, if no active same-polarity record exists, we set $z_i=1$. Otherwise, let
\[
    j^\star
    =
    \arg\max_{j:z_j=1,\;p_j=p_i}
    \operatorname{sim}(i,j).
\]
If $\operatorname{sim}(i,j^\star)<\theta$, where $\theta$ is set by the corresponding experiment configuration, the new argument remains active. If the similarity exceeds the threshold, only the stronger of the two near-duplicate records remains active:
\[
    (z_i,z_{j^\star}) =
    \begin{cases}
        (1,0), & s_i > s_{j^\star},\\
        (0,1), & s_i \leq s_{j^\star}.
    \end{cases}
\]
The archived record remains stored, but it is excluded from the active set used for belief updating and retrieval.
Generated-agent thresholds are reported in Tab.~\ref{tab:hyperparams}, and DEBATE replay uses $\theta=0.85$.

\subsection{Initialization and update modes}
\label{sec:update-semantics}
The implementation stores each structured record as an \texttt{ArgumentRecord}. Its \texttt{credence\_relevant} field corresponds to the active flag $z_i$.
In the seed-derived mode (\texttt{active\_seed\_v1}), Eq.~\ref{eq:belief-update} recomputes the state from active seed and later-message records. In the explicit-prior mode (\texttt{active\_explicit\_prior\_v1}), a fixed initial stance $S_0$ instead supplies log odds $\ell(S_0)=\ln((1+S_0)/(1-S_0))$, and only active non-seed contributions are added. Seed records remain retrievable but are excluded from the numerical sum to avoid counting the prior twice. Therefore, $a$ does not affect the numerical state in this mode.

The submitted incremental implementation (\texttt{incremental\_v1}) accumulated newly admitted contributions without recomputing the active set. On replacement, it changed the active flags but neither subtracted the old contribution nor added the replacement contribution. Archiving therefore did not undo previously applied stance movement. A configured initial target could also overwrite the state after seeding. These historical trajectories cannot generally be reconstructed from the final active set and must be distinguished from active-set recomputation.

\section{Common-Evaluator Baseline Comparison}
\label{sec:common-evaluator-details}

\paragraph{Matched protocol.}
The new comparison comprises five preregistered topics, ten matched seed blocks per topic, and three methods: BE, Prompt self-update, and RAG+self-update. All 150 debates run for 15 rounds from Pro/Con targets $+0.75$ and $-0.75$, using GPT-5.4-mini for generation at temperature $0.7$ with a shared response shell and response budget. Recent dialogue contains four messages. BE and RAG retrieve up to ten active records. BE uses uptake $u=0.2$ and an explicit initial prior plus active non-seed evidence (\texttt{active\_explicit\_prior\_v1}). Seeds remain retrievable but do not enter the numerical sum. The configured $a=0.4$ therefore has no numerical effect in this comparison. These runs are distinct from the earlier three-trial diagnostics.

\paragraph{Common outcome.}
A frozen GPT-5.4-mini evaluator, at temperature $0.0$ with condition labels removed, scores every method's generated text on $[-1,1]$. The primary outcome is the absolute difference between final Pro and Con scores, ranging from 0 to 2. Generator and evaluator share a model family, so the comparison controls measurement consistency but does not establish evaluator-independent validity.

\begin{table}[H]
\centering
\caption{Final externally judged disagreement by topic in the common-evaluator comparison. Each topic has ten matched blocks per method. Smaller values indicate closer expressed positions.}
\label{tab:common-evaluator-topics}
\begin{tabular}{lrrr}
\toprule
Topic & BE & Prompt self-update & RAG+self-update \\
\midrule
Social media & 0.174 & 1.450 & 1.246 \\
Journalism & 0.813 & 1.018 & 1.284 \\
Compulsory voting & 0.546 & 0.993 & 1.003 \\
Atheism & 0.522 & 1.135 & 0.771 \\
Libertarianism & 0.586 & 0.983 & 0.787 \\
\midrule
Equal-topic mean & 0.528 & 1.116 & 1.018 \\
\bottomrule
\end{tabular}
\end{table}

\paragraph{Uncertainty and sensitivity.}
The experimental unit is a matched topic--seed block. Mean final gaps and 95\% intervals are $0.53$ [$0.31,0.75$] for BE, $1.12$ [$0.97,1.31$] for Prompt self-update, and $1.02$ [$0.76,1.24$] for RAG+self-update. Condition-level 95\% intervals use 10,000 hierarchical bootstrap draws, resampling topics and then paired seed blocks. The two planned BE-minus-baseline contrasts receive simultaneous 95\% intervals from a separate post-hoc multiplicity audit using 100,000 shared hierarchical draws and a maximum absolute studentized-deviation statistic. The BE-minus-Prompt difference is $-0.59$ [$-0.99,-0.19$], and BE minus RAG is $-0.49$ [$-0.86,-0.12$]. The corresponding Bonferroni percentile intervals are $[-1.033,-0.214]$ for BE minus Prompt and $[-0.884,-0.111]$ for BE minus RAG. Leave-one-topic-out estimates range from $-0.683$ to $-0.416$ and from $-0.562$ to $-0.345$, respectively. An exhaustive sign-flip check over the $2^5=32$ topic-sign assignments gives two-sided $p=0.0625$ for each contrast, also after the joint adjustment. This coarse five-topic check limits topic-general claims despite the negative bootstrap intervals. Smaller disagreement alone does not establish greater truthfulness, human realism, neutrality, better reasoning, or preferable consensus.

\section{Earlier Five-Topic Prompt-Based Check}
\label{sec:five-topic-prompt-based-check}

The five-topic mechanism check broadens the prompt-based comparison beyond the main atheism example (Tab.~\ref{tab:five-topic-prompt-based-check}). Each run pairs a Pro agent and a Con agent for 15 rounds, with fixed initial stance anchors $+0.75$ and $-0.75$, respectively. We run three independent trials per topic--setup pair and summarise the final stance dynamics across five topics: social media, journalism, compulsory voting, atheism, and libertarianism.

The four Belief Engine rows use the same memory, retrieval, and update profile across topics, with evidence uptake $u=0.2$ and prior anchoring $a=0.4$, and differ only in the base language model generating debate turns. The two non-BE baselines use GPT-5.4-mini. In Prompt-only, agents have no retrieval memory or Belief Engine and are instead prompted to update their stance after opponent turns. In RAG, agents receive retrieved argument memory and prompt-level self-updating, but still do not use the Belief Engine stance update.

The five-topic check keeps the same measurement caveat as Fig.~\ref{fig:mechanism-and-calibration}. BE rows report internal stance, whereas baseline rows report external judge scores because those systems do not expose $S$. The external judge is a calibrated proxy for expressed stance and cannot replace an auditable internal trajectory.

\begin{table}[H]
\centering
\caption{Five-topic prompt-based mechanism check over 15-round Pro/Con debates. Each row averages three trials for one topic and setup. BE variants share the same retrieval and update profile, with $u=0.2$ and $a=0.4$, and differ only in the base debate model. Prompt-only and RAG are GPT-5.4-mini non-BE baselines. BE rows use internal stance, and prompt-based baselines use external judge scores.}

\label{tab:five-topic-prompt-based-check}
\scriptsize
\setlength{\tabcolsep}{3pt}
\resizebox{\linewidth}{!}{%
\begin{tabular}{lllrrrrrrrrr}
\toprule
Topic & Setup & Measure & $n$ & Final Pro & Final Con & \textbar{}Final gap\textbar{} & Gap reduction & Mean abs.\ shift & Centre shift & Crossing rate \\
\midrule
Social media & BE GPT-4o-mini & BE internal & 3 & 0.36 & 0.06 & 0.38 & 1.12 & 0.60 & 0.60 & 0.33 \\
Social media & BE GPT-5.4-mini & BE internal & 3 & 0.84 & 0.95 & 0.14 & 1.36 & 0.96 & 0.81 & 0.67 \\
Social media & BE Qwen 3.5 9B & BE internal & 3 & -0.02 & -0.17 & 0.18 & 1.32 & 0.68 & 0.68 & 0.33 \\
Social media & BE Gemma 4 E4B & BE internal & 3 & -0.16 & 0.73 & 0.89 & 0.61 & 1.20 & 1.20 & 1.00 \\
Social media & Prompt-only & External judge & 3 & -0.03 & 0.05 & 0.31 & 1.19 & 0.79 & 0.79 & 0.33 \\
Social media & RAG & External judge & 3 & -0.11 & 0.34 & 0.68 & 0.82 & 0.98 & 0.98 & 0.67 \\
Journalism & BE GPT-4o-mini & BE internal & 3 & 0.02 & -0.08 & 0.14 & 1.36 & 0.70 & 0.70 & 0.33 \\
Journalism & BE GPT-5.4-mini & BE internal & 3 & 0.11 & 0.33 & 0.22 & 1.28 & 0.86 & 0.86 & 1.00 \\
Journalism & BE Qwen 3.5 9B & BE internal & 3 & 0.58 & 0.27 & 0.39 & 1.11 & 0.60 & 0.60 & 0.33 \\
Journalism & BE Gemma 4 E4B & BE internal & 3 & 0.36 & 0.39 & 0.15 & 1.35 & 0.76 & 0.76 & 0.33 \\
Journalism & Prompt-only & External judge & 3 & 0.34 & -0.73 & 1.07 & 0.43 & 0.24 & 0.22 & 0.00 \\
Journalism & RAG & External judge & 3 & 0.27 & -0.58 & 0.85 & 0.65 & 0.34 & 0.33 & 0.00 \\
Compulsory voting & BE GPT-4o-mini & BE internal & 3 & 0.05 & -0.15 & 0.26 & 1.24 & 0.65 & 0.65 & 0.33 \\
Compulsory voting & BE GPT-5.4-mini & BE internal & 3 & -0.37 & -0.08 & 0.29 & 1.21 & 0.90 & 0.90 & 1.00 \\
Compulsory voting & BE Qwen 3.5 9B & BE internal & 3 & 0.48 & 0.24 & 0.27 & 1.23 & 0.64 & 0.63 & 0.33 \\
Compulsory voting & BE Gemma 4 E4B & BE internal & 3 & -0.17 & 0.06 & 0.22 & 1.28 & 0.86 & 0.86 & 1.00 \\
Compulsory voting & Prompt-only & External judge & 3 & 0.27 & -0.34 & 0.61 & 0.89 & 0.45 & 0.45 & 0.00 \\
Compulsory voting & RAG & External judge & 3 & 0.22 & -0.30 & 0.53 & 0.97 & 0.49 & 0.49 & 0.00 \\
Atheism & BE GPT-4o-mini & BE internal & 3 & 0.16 & -0.34 & 0.50 & 1.00 & 0.50 & 0.50 & 0.00 \\
Atheism & BE GPT-5.4-mini & BE internal & 3 & 0.17 & 0.40 & 0.23 & 1.27 & 0.86 & 0.86 & 1.00 \\
Atheism & BE Qwen 3.5 9B & BE internal & 3 & 0.09 & -0.12 & 0.22 & 1.28 & 0.65 & 0.64 & 0.00 \\
Atheism & BE Gemma 4 E4B & BE internal & 3 & 0.74 & 0.76 & 0.07 & 1.43 & 0.84 & 0.76 & 0.67 \\
Atheism & Prompt-only & External judge & 3 & 0.23 & -0.40 & 0.63 & 0.87 & 0.45 & 0.44 & 0.00 \\
Atheism & RAG & External judge & 3 & 0.23 & -0.62 & 0.85 & 0.65 & 0.37 & 0.33 & 0.00 \\
Libertarianism & BE GPT-4o-mini & BE internal & 3 & -0.26 & -0.04 & 0.23 & 1.27 & 0.86 & 0.86 & 0.67 \\
Libertarianism & BE GPT-5.4-mini & BE internal & 3 & -0.36 & 0.40 & 0.76 & 0.74 & 1.13 & 1.13 & 1.00 \\
Libertarianism & BE Qwen 3.5 9B & BE internal & 3 & 0.60 & 0.09 & 0.51 & 0.99 & 0.56 & 0.49 & 0.00 \\
Libertarianism & BE Gemma 4 E4B & BE internal & 3 & 0.39 & 0.90 & 0.50 & 1.00 & 1.02 & 1.00 & 1.00 \\
Libertarianism & Prompt-only & External judge & 3 & 0.60 & -0.51 & 1.11 & 0.39 & 0.25 & 0.19 & 0.00 \\
Libertarianism & RAG & External judge & 3 & 0.41 & -0.49 & 0.90 & 0.60 & 0.32 & 0.30 & 0.00 \\
\bottomrule
\end{tabular}%
}
\end{table}

\begin{figure}[H]
    \centering
    \includegraphics[width=\linewidth]{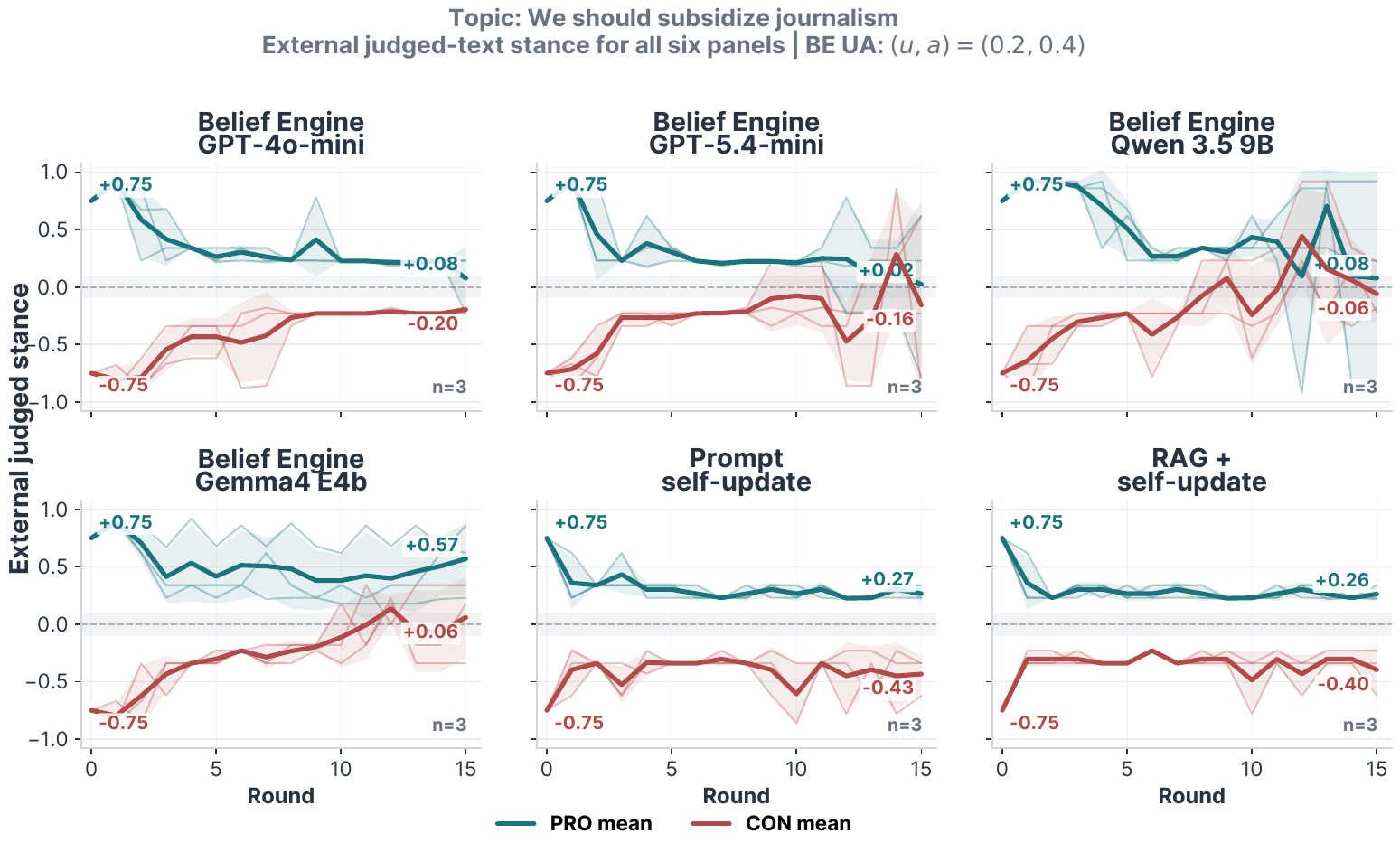}
    \caption{External-judge trajectory check on the journalism topic, ``We should subsidize journalism.'' All panels use the same judged-text stance metric, from $-1$ strongly con to $+1$ strongly pro. Thick lines show mean judged stance across three trials for Pro-side and Con-side agents. Faint lines show individual trials and bands show across-trial variation. This figure is an illustrative comparability check.}
    \label{fig:external-judge-journalism}
\end{figure}

As an illustrative measurement-controlled check, we rescored the generated journalism debates with the same external stance judge for all setups (Fig.~\ref{fig:external-judge-journalism}). This removes the mixed-measurement caveat in Tab.~\ref{tab:five-topic-prompt-based-check}, since all setups are now scored the same way. The figure should be read qualitatively, since it covers one topic with three trials per setup, but it supports the same pattern. Most BE variants move both sides closer to the centre under the external judge, while Prompt-only and RAG leave a larger final gap.

\clearpage
\section{Alignment Between Maintained and Expressed Stance}
\label{sec:stance-alignment-details}

\paragraph{Controlled protocol.}
We extend the original compulsory-voting check to ten propositions. For each proposition, we set the maintained stance to 21 positions from $-1$ to $+1$ in steps of $0.1$ and generate three responses at each position. This produces 630 response--judgment pairs, all completed successfully. Task order is randomized with seed 20260727. The generator is GPT-4o-mini (\texttt{gpt-4o-mini-2024-07-18}) at temperature $0.7$. It uses the production SubjectAgent prompt and the Belief Engine's stance description, with no retrieved arguments or conversation history. GPT-5.4-mini judges the proposition and generated response at temperature $0$, using a frozen evaluator prompt. The judge does not receive the internal score or the generator's stance instruction.

\paragraph{Estimates and uncertainty.}
We average the three external scores within each topic--stance cell before computing the primary statistics. The analysis therefore contains 210 cell means, with 21 per topic. This avoids treating repeated generations at the same position as additional stance positions. Pearson correlation measures how closely the scores track the assigned stance, while the OLS slope, fitted with an intercept, measures their relative scale. A slope below one indicates compression on average. We obtain pooled 95\% confidence intervals from 5,000 bootstrap samples that resample whole topics with replacement, retaining all 21 cell means within each selected topic, using seed 20260727.

Across the ten topics, the pooled correlation is $r=0.959$ (95\% CI $[0.952,0.964]$) and the slope is $0.928$ (95\% CI $[0.905,0.950]$). The mean absolute error is $0.144$ and the root mean squared error is $0.173$. Table~\ref{tab:stance-alignment-topics} gives the per-topic estimates. Figure~\ref{fig:stance-alignment-topics} shows the response patterns and variation across repeated generations.

\begin{table}[htbp]
\centering
\caption{Maintained versus externally judged stance across ten propositions. Each topic has 21 stance positions and 63 generated responses. Statistics use the 21 cell means. The final column recomputes the pooled correlation after excluding the named topic. MAE is mean absolute error on the $[-1,1]$ stance scale.}
\label{tab:stance-alignment-topics}
\setlength{\tabcolsep}{5pt}
\renewcommand{\arraystretch}{1.15}
\begin{tabular}{@{}>{\raggedright\arraybackslash}p{0.49\linewidth}rrrr@{}}
\toprule
Proposition & $r$ & Slope & MAE & Excluded $r$ \\
\midrule
Entrapment should be legalized & 0.951 & 0.912 & 0.167 & 0.959 \\
Homeopathy brings more harm than good & 0.960 & 0.922 & 0.141 & 0.958 \\
Social media brings more harm than good & 0.970 & 0.985 & 0.134 & 0.957 \\
We should adopt a zero-tolerance policy in schools & 0.968 & 0.968 & 0.134 & 0.957 \\
We should adopt an austerity regime & 0.961 & 0.939 & 0.136 & 0.958 \\
We should adopt atheism & 0.934 & 0.849 & 0.169 & 0.961 \\
We should adopt libertarianism & 0.963 & 0.948 & 0.145 & 0.958 \\
We should introduce compulsory voting & 0.950 & 0.892 & 0.145 & 0.959 \\
We should legalize sex selection & 0.964 & 0.944 & 0.136 & 0.958 \\
We should subsidize journalism & 0.966 & 0.924 & 0.133 & 0.958 \\
\bottomrule
\end{tabular}
\end{table}

\paragraph{Robustness and interpretation.}
Per-topic correlations range from $0.934$ to $0.970$. Removing each topic's mean from both variables before pooling gives $r=0.959$, and the Fisher-$z$ macro-average of the ten topic correlations is $0.960$. Leaving out one topic at a time gives pooled correlations between $0.957$ and $0.961$, so the aggregate result is not driven by a single proposition. Slopes range from $0.849$ to $0.985$, showing that strong correlation can coexist with compression. These checks establish alignment between controlled stance and automated judgments of generated text under this protocol. They do not establish agreement with human readers or validate the scalar state as a model of human belief. This expanded study is separate from the earlier 201-position compulsory-voting check ($r=0.967$, slope approximately $0.86$).

\begin{figure}[p]
\centering
\resizebox{\linewidth}{!}{\input{figures/stance_alignment_ten_topics.tex}}
\caption{Maintained and expressed stance for all ten propositions. Each teal point is the mean external score of three responses at one assigned stance. Error bars show one standard deviation across those responses, not confidence intervals. Red lines are OLS fits with intercepts, and dashed diagonal lines show exact agreement. The shaded band marks external scores between $-0.1$ and $+0.1$. All panels use the same scales.}
\label{fig:stance-alignment-topics}
\end{figure}
\clearpage

\section{Judgement Layer Validation Details}
\label{sec:judgement-validation-details}

We evaluate argument extraction, polarity, and strength scoring using the Argument Quality Ranking (AQR) dataset~\citep{Gretz2019argumentquality}. AQR contains clean single arguments with crowd-derived pro/con labels and crowd-rated argument quality scores. These labels let us check whether the system recovers an argument's direction and whether its strength scores track human-rated quality. They do not establish how much an argument should change a person's stance.

For the extraction audit, we use 100 single-argument inputs. The extractor returned at least one claim for 97, exactly one claim for 77, and several subclaims for 20. Among the 97 inputs that yielded an argument, the strongest extracted claim had the correct polarity in 94 (96.9\% accuracy, 0.969 macro F1). These results support polarity recovery on clean, in-domain arguments.

We also check strength scores against crowd-rated quality, using the mean of AQR's WA and MACE-P ratings as the target. On all 100 inputs in this audit, classifier scores correlated with that target at $r=0.720$. LLM-assigned strengths showed little correlation on the 97 inputs that yielded an extracted claim ($r=0.089$). These summaries therefore have different denominators. The classifier was trained on the same dataset family, so its result describes agreement on this in-domain sample. Table~\ref{tab:aqr-judgement} reports the sample size for each metric.

\begin{table}[H]
\centering
\caption{Judgement-layer diagnostics on AQR. Extraction rates use all 100 inputs. Polarity and LLM-strength results use the 97 inputs that yielded an argument. Classifier strength uses all 100 inputs from the same audit. Correlations measure agreement with crowd-rated quality, not stance-update magnitude.}
\label{tab:aqr-judgement}
\small
\setlength{\tabcolsep}{4pt}
\renewcommand{\arraystretch}{1.12}
\begin{tabular}{@{}>{\raggedright\arraybackslash}p{0.37\linewidth}>{\centering\arraybackslash}p{0.10\linewidth}>{\centering\arraybackslash}p{0.10\linewidth}>{\raggedright\arraybackslash}p{0.36\linewidth}@{}}
\toprule
\textbf{Metric} & \textbf{$N$} & \textbf{Value} & \textbf{Interpretation} \\
\midrule
Extraction rate & 100 & 0.970 & At least one claim extracted. \\
\addlinespace[3pt]
Exact-one extraction rate & 100 & 0.770 & Exactly one claim extracted. \\
\addlinespace[3pt]
Over-split rate & 100 & 0.200 & A single argument split into several claims. \\
\addlinespace[3pt]
Strongest-claim polarity accuracy & 97 & 0.969 & Extracted direction matches the pro/con label. \\
\addlinespace[3pt]
Strongest-claim macro F1 & 97 & 0.969 & F1 averaged across the two polarity labels. \\
\addlinespace[3pt]
LLM strength vs.\ AQR quality ($r$) & 97 & 0.089 & Little correlation with crowd-rated quality. \\
\addlinespace[3pt]
Classifier strength vs.\ AQR quality ($r$) & 100 & 0.720 & Agreement with crowd-rated quality on the 100-input audit sample. \\
\bottomrule
\end{tabular}
\end{table}

\section{DEBATE Replay Protocol and Robustness}
\label{sec:debate-replay-protocol}

DEBATE contains 2{,}792 U.S.-based participants in four-person discussions over 107 controversial topics: participants privately report an initial opinion and justification, exchange round-wise tweet-like posts and dyadic chat messages, and privately report a final opinion. Our replay uses 2{,}495 quality-filtered participant trajectories with usable pre/post stance reports and processable directed-exposure histories. Six-point Likert responses use fixed log-odds anchors $(-3,-1.5,-0.5,0.5,1.5,3)$, mapped to stance by $S=2\sigma(L)-1$. For each participant, the mapped initial Likert stance initialises the prior. Prior anchoring $a$ scales that initial log-odds state, and arguments extracted from directed partner tweets and received chat messages are replayed chronologically under evidence uptake $u$. The replay evaluates belief updating under a fixed received-evidence stream, without generating new responses. It uses the same log-odds evidence contributions as Eq.~\ref{eq:belief-update}, with the prior supplied by the participant's initial stance rather than seed arguments.

The main replay uses directed received exposure, human-like uptake and vector deduplication, classifier-based argument strength, five held-out group folds, and seed 42. Appendix~\ref{sec:debate-fold-robustness} compares these results with a split that holds out entire topics. For each fold, $(u,a)$ is selected on training groups from the grid below and evaluated on held-out groups by RMSE against the participant's mapped final stance.

The net-evidence linear baseline is fit on the same group-held-out replay setup. For participant $j$ in fold $f$, it predicts
\[
    \hat S_{j,\mathrm{final}}
    =
    S_{j,\mathrm{initial}}
    +
    \beta_f E_j,
\]
where $E_j=\sum_{i\in\mathcal{R}_j} p_i s_i$ is the signed net evidence over accepted received records $\mathcal{R}_j$, after the same quality filtering, human-like uptake/deduplication filter, and classifier-strength scoring used by BE. We fit one scalar coefficient $\beta_f$ on training folds only, then evaluate on the held-out fold. The five fitted coefficients were small: $0.0083$, $0.0105$, $0.0080$, $0.0089$, and $0.0082$.

\begin{center}
\small
\begin{tabular}{lp{0.76\linewidth}}
$u$ grid & $0.005$, $0.01$, $0.02$, $0.035$, $0.05$, $0.075$, $0.1$, $0.15$, $0.2$, $0.3$, $0.4$, $0.6$, $0.8$ \\
$a$ grid & $0.02$, $0.05$, $0.1$, $0.15$, $0.2$, $0.3$, $0.4$, $0.5$, $0.6$, $0.8$, $1.0$, $1.2$, $1.5$ \\
\end{tabular}
\end{center}

\begin{table}[H]
\centering
\caption{Evidence-alignment grouping used for DEBATE replay calibration. These are outcome-conditioned diagnostic groups used to analyse replay behaviour.}
\label{tab:debate-replay-groups}
\small
\setlength{\tabcolsep}{4pt}
\begin{tabular}{p{0.24\linewidth}rp{0.58\linewidth}}
\toprule
Group & $N$ & Definition and use \\
\midrule
All participants & 2,495 & Full quality-filtered replay population used for the pooled default calibration. \\
Evidence-aligned movers & 542 & Non-stable participants whose observed movement has the same sign as net accepted evidence from received messages. \\
Evidence-opposed movers & 400 & Non-stable participants whose observed movement has the opposite sign from net accepted evidence from received messages. \\
Weak-signal movers & 174 & Non-stable participants with weak or mixed accepted-evidence pressure; retained in the pooled calibration but excluded from the three headline profiles. \\
Stable participants & 1,379 & Participants with no mapped Likert movement; this combines stable-low-pressure and stable-despite-pressure cases in the calibration code. \\
\bottomrule
\end{tabular}
\end{table}

Table~\ref{tab:debate-main-rmse} in the main text reports the replay errors. The three participant subgroups are defined using observed final opinion change, so their results describe how well the model reconstructs known outcomes. They do not establish that the appropriate profile can be predicted in advance. The 174 weak-signal movers are included in the All participants row but are not reported as a separate subgroup in that table.

\subsection{Same final stance, different trajectories}
\label{sec:endpoint-paths}

A final stance score does not show how the update rule responded to evidence along the way. We therefore asked whether two settings of evidence uptake $u$ and prior anchoring $a$ could reach the same final stance while following different paths. For each participant in DEBATE, we kept the accepted evidence and its order fixed, then searched for two settings whose final stance scores matched. This comparison changes how the rule weighs the prior and the evidence, while keeping the evidence itself the same.

\paragraph{What we found.}
Among 2,441 eligible participant records, we found matching final stances for 958. In 869 of those matches, the two stance scores differed by at least $0.20$ after an earlier evidence item, even though they ended at the same value. These cases account for 35.60\% of all eligible records and 90.71\% of the detected matches (Table~\ref{tab:endpoint-paths}). Thus, comparing only the final scores would miss differences in how the two settings responded during the replay.

\begin{table}[H]
\centering
\caption{How often different settings reach the same final stance after receiving the same evidence. The first two rows count cases among all 2,441 eligible records. The last row counts earlier differences among the 958 records with matching final stances.}
\label{tab:endpoint-paths}
\small
\begin{tabular}{lrrr}
\toprule
Outcome & Count & Rate & 95\% CI \\
\midrule
Matching final stance & 958/2,441 & 39.25\% & [36.38, 42.14]\% \\
Matching final stance and earlier gap $\geq0.20$ & 869/2,441 & 35.60\% & [32.75, 38.46]\% \\
Earlier gap $\geq0.20$ among matches & 869/958 & 90.71\% & [87.64, 93.52]\% \\
\bottomrule
\end{tabular}
\end{table}

\begin{figure}[H]
\centering
\includegraphics[width=\linewidth]{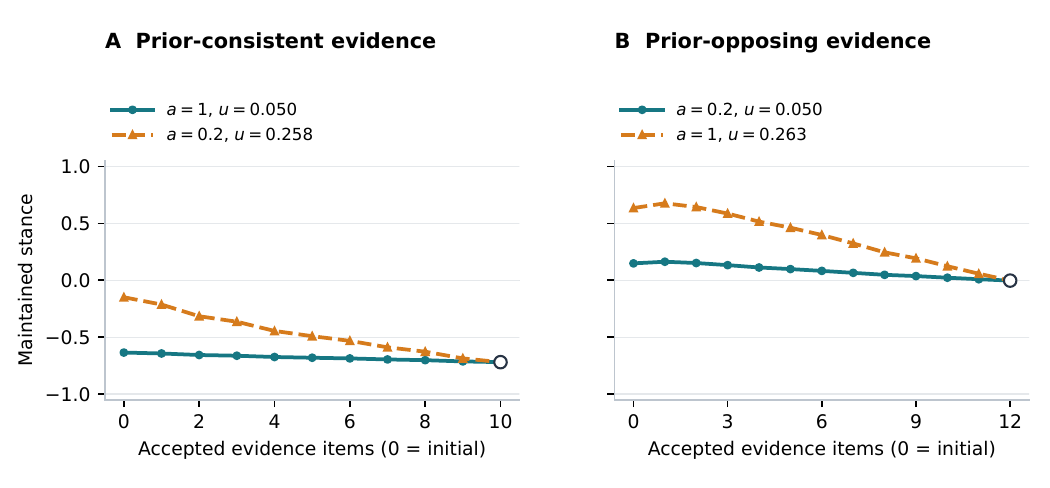}
\caption{Different paths to the same final stance. Each panel replays one participant's accepted evidence under two uptake--anchoring settings. Teal curves use lower uptake and orange curves use higher uptake. Giving more weight to the evidence reinforces the initial position in A and opposes it in B. Open circles mark the shared final stance. The largest differences after an intermediate evidence item are $0.4301$ (A) and $0.5136$ (B). Uptake values in the legends are rounded.}
\label{fig:endpoint-paths}
\end{figure}

Figure~\ref{fig:endpoint-paths} illustrates the comparison with two records. The lower-uptake setting changes relatively little, while the higher-uptake setting moves further as evidence arrives. Both finish at the same stance. Their starting points differ because anchoring gives different weights to the same participant's initial opinion.

\paragraph{How we constructed the comparisons.}
We used the DEBATE replay formulation, where the participant's observed initial stance $S_{\mathrm{init}}$ supplies the prior. Anchoring $a$ scales its log odds, and uptake $u$ controls the contribution of each accepted evidence item:
\[
L_t(a,u)=a\ell(S_{\mathrm{init}})+\sum_{i=1}^{t}p_i\ln(1+u s_i),
\qquad \ell(S)=\ln\frac{1+S}{1-S},
\qquad S_t(a,u)=2\sigma(L_t(a,u))-1.
\]
Here $t$ counts accepted evidence items, $p_i$ is an item's polarity, and $s_i$ is its strength. Both settings use the same items, polarities, and strengths. At $t=0$, the evidence sum is empty, so the maintained stance depends on the initial opinion and its anchoring weight. This analysis uses an observed initial opinion as the prior, rather than generated seed arguments.

The two anchoring settings were $a=0.2$ and $a=1.0$. We fixed uptake at $u=0.05$ for one setting and searched over $u\in[0.25,0.8]$ for the other. Which setting received the stronger anchor depended on whether increasing uptake reinforced or opposed the initial position. To determine this, we computed the total evidence contribution $E(u)=\sum_{i=1}^{T}p_i\ln(1+u s_i)$ and the change $E(0.8)-E(0.05)$. When this change had the same sign as $\ell(S_{\mathrm{init}})$, we compared $(a,u)=(1.0,0.05)$ with $(0.2,u^*)$. When the signs differed, we compared $(0.2,0.05)$ with $(1.0,u^*)$. In each case, $u^*$ was the smallest uptake value found that produced the same final stance as the fixed-uptake setting. The parameter comparison, eligibility criteria, and example selection did not use participants' observed final opinions.

\paragraph{Numerical search and eligible records.}
The solver searched for uptake values where the difference between the two final stance scores was zero. We divided the search range into 8,192 intervals and applied Brent's method wherever that difference changed sign. Exact zeros at scan points were also retained. Absolute and relative solver tolerances were $5\times10^{-15}$ and $10^{-14}$, respectively. We counted a match when final stance scores differed by at most $10^{-10}$. The largest observed difference among matches was $6.48\times10^{-15}$.

Of the 2,495 source records, 53 had fewer than three accepted evidence items and one had a reference final stance at or beyond $\pm0.95$. No records met the exclusion criterion for negligible change in evidence contribution, $|E(0.8)-E(0.05)|\leq10^{-14}$. The remaining 2,441 records covered 694 discussion groups and 107 topics. We retained the 1,483 records with no detected match when calculating the rates among all eligible records.

\paragraph{Path differences and uncertainty.}
For each match, we measured the largest absolute stance difference after an intermediate evidence item, $\max_{1\leq t<T}|S_t^{\mathrm{reference}}-S_t^{\mathrm{matched}}|$. This excludes the initial and final points. The median of these maximum differences was $0.4405$ (95\% CI $[0.4237,0.4544]$). To select the examples in Figure~\ref{fig:endpoint-paths}, we restricted attention to matches with a difference of at least $0.20$. Within each evidence direction, we chose the record closest to that subset's median difference, breaking ties by participant identifier.

We computed percentile confidence intervals from 100,000 bootstrap draws, sampling topics and then intact discussion groups within each sampled topic (seed 2026072502). The share of eligible records with both matching final stances and an earlier gap of at least $0.20$ was similar when increasing uptake reinforced or opposed the prior (35.35\% and 35.92\%). Leaving out each topic in turn gave estimates from 35.31\% to 35.93\%. Sampling groups alone gave a 95\% interval of [33.60, 37.62]\%. Keeping topics fixed and resampling groups within them gave [33.82, 37.39]\%.

\paragraph{What this analysis establishes.}
The result shows that different uptake--anchoring settings can reach the same final stance through different intermediate states. Recording the full trajectory makes those differences visible. Because we hold the received evidence fixed, this test does not establish what would happen in a generated conversation, where changing a reply can also change the evidence received next. It also does not identify human cognitive parameters or show that matching internal stances produce identical replies. Finally, the numerical search is finite. It can miss a zero that does not change sign or multiple zeros within one scan interval, so failure to find a match does not establish that none exists.

\subsection{Robustness across discussion groups and topics}
\label{sec:debate-fold-robustness}

We checked whether the replay results were consistent when holding out entire discussion groups or entire topics. For each split, we computed one held-out RMSE for each of five outer folds. Table~\ref{tab:debate-fold-robustness} reports the mean error and comparisons with the prior-only and pooled global replay references. The 95\% confidence intervals are $t$ intervals across the five folds. They summarise variation across folds and do not estimate uncertainty at the participant level.

\begin{table}[H]
\centering
\caption{Replay errors across five held-out folds. Differences are the RMSE of the group-specific uptake--anchoring settings minus that of each reference model. Negative differences favour the group-specific settings.}
\label{tab:debate-fold-robustness}
\scriptsize
\setlength{\tabcolsep}{4pt}
\begin{tabular}{lcccc}
\toprule
Split & \shortstack{Fold mean\\RMSE} & 95\% CI &
\shortstack{$\Delta$ vs\\prior-only} & \shortstack{$\Delta$ vs\\global} \\
\midrule
Group-held-out & 0.365 & [0.337, 0.393] & -0.123 [-0.145, -0.102] & -0.100 [-0.112, -0.087] \\
Topic-held-out & 0.366 & [0.356, 0.376] & -0.122 [-0.152, -0.092] & -0.099 [-0.118, -0.081] \\
\bottomrule
\end{tabular}
\end{table}

The group-specific settings have lower RMSE than both references in every fold under both splits, as well as lower aggregate RMSE. However, the participant groups are defined using observed opinion change. This check therefore assesses how well the settings reconstruct known outcomes. It does not show that we can select a participant's profile before their final opinion is known.

\section{Implementation and Reproducibility Details}
\label{sec:repro-prompts-models}

This appendix brings together the experiment settings, model roles, prompts, and release information. The detailed baseline and human-replay protocols remain in Appendices~\ref{sec:common-evaluator-details} and \ref{sec:debate-replay-protocol}.

\subsection{Experiment settings and update modes}
\label{sec:experimental-hyperparameters}

Table~\ref{tab:hyperparams} reports settings for the earlier generated-agent diagnostics, including the parameter sweeps and profile demonstrations. These settings do not apply to every experiment. The common-evaluator comparison uses the separate settings in Appendix~\ref{sec:common-evaluator-details}, including retrieval of up to ten records per turn. The DEBATE replay uses the parameter grid and held-out splits in Appendix~\ref{sec:debate-replay-protocol}.

The prior also enters the update differently across protocols. In the common-evaluator comparison, BE starts from a fixed stance and recomputes contributions from active non-seed records. Seed records remain available for retrieval, but anchoring does not change the numerical state. In DEBATE replay, anchoring instead scales the participant's initial log odds before received evidence is added, as written in Appendix~\ref{sec:endpoint-paths}. Appendix~\ref{sec:update-semantics} distinguishes the seed-derived and explicit-prior modes from the historical incremental implementation, where archiving a record did not undo its earlier contribution.

\begin{table}[H]
\centering
\caption{Hyperparameters and scoring settings for the earlier generated-agent diagnostics. The common-evaluator comparison uses the settings in Appendix~\ref{sec:common-evaluator-details}. The Belief Engine controls are evidence uptake $u$ and prior anchoring $a$. Values are fixed unless a row explicitly states a sweep or profile-specific setting.}
\label{tab:hyperparams}
\scriptsize
\setlength{\tabcolsep}{4pt}
\begin{tabular}{p{0.21\linewidth}p{0.21\linewidth}p{0.32\linewidth}p{0.18\linewidth}}
\toprule
\textbf{Setting} & \textbf{Symbol / config key} & \textbf{Value(s)} & \textbf{Used in} \\
\midrule
Evidence uptake & \shortstack[l]{$u$\\\texttt{update\_sensitivity}} &
Sweep: $\{0.2,0.4,0.6,0.8,1.0\}$; fixed prompt-baseline BE rows: $0.2$; Open-minded: $0.40$; Stubborn: $0.10$ &
Parameter sweeps, prompt-baseline comparison, profile grids \\
Prior anchoring & \shortstack[l]{$a$\\\texttt{anchor\_sensitivity}} &
Sweep: $\{0.2,0.4,0.6,0.8,1.0\}$; fixed prompt-baseline BE rows: $0.4$; Open-minded: $0.20$; Stubborn: $0.80$ &
Parameter sweeps, prompt-baseline comparison, profile grids \\
Confirmation asymmetry & \shortstack[l]{$B$\\\texttt{confirmation\_bias}} &
$0.0$ &
All reported generated-agent runs \\
Argument deduplication & \shortstack[l]{$\theta$\\\texttt{argument\_similarity}\\\texttt{\_threshold}} &
$0.80$ in model sweeps; $0.60$ in matched-baseline and Open-minded/Stubborn profile runs &
Memory admission and replacement \\
Self-argument deduplication & \shortstack[l]{$\theta_{\mathrm{self}}$\\\texttt{self\_similarity}\\\texttt{\_threshold}} &
$0.50$ in model sweeps; $0.45$ in matched-baseline and profile runs &
Self-generated argument filtering \\
Seed arguments & \shortstack[l]{$n$\\\texttt{seeding\_limit}} &
Limit $10$ per side; $8$ admitted in single-agent sweeps &
All generated-agent runs \\
Retrieved context & $k$ &
At most $5$ active arguments per turn &
BE and RAG conditions \\
Debate horizon & \shortstack[l]{$R$\\\texttt{rounds}} &
$15$ rounds &
All generated-agent runs \\
Initial stance target & $S_0$ &
Uptake sweeps: target $+0.99$, realised $+0.928$; anchoring sweeps: seed-derived, varying with $a$; two-agent comparisons and profile grids: $(+0.75,-0.75)$ &
Initialisation \\
Trials & -- &
$5$ per sweep setting; $3$ per matched-baseline topic/setup and profile-pairing cell &
Aggregation \\
Agent generation temperature & $T_{\mathrm{agent}}$ &
$0.7$ &
Generated utterances \\
External stance judge & -- &
Scores text in $[-1,1]$ at temperature $0.0$ when no internal stance is available &
Prompt baselines \\
\bottomrule
\end{tabular}
\end{table}

\subsection{Model identifiers}
\label{sec:model-identifiers}

\begin{table}[H]
\centering
\caption{Model identifiers and evaluator roles used in the reported experiments. Release locations and reproduction instructions are given in Appendix~\ref{sec:code-release-plan}.}
\label{tab:model-identifiers}
\scriptsize
\setlength{\tabcolsep}{4pt}
\begin{tabular}{p{0.30\linewidth}p{0.26\linewidth}p{0.36\linewidth}}
\toprule
\textbf{Use} & \textbf{Model family / identifier} & \textbf{Reporting note} \\
\midrule
Generated BE sweeps & \texttt{gpt-4o-mini}; Qwen 3.5 9B; Gemma 4 E4B & Used to test whether the same uptake/anchoring controls behave consistently across hosted and local model backends. \\
Matched generated-agent and prompt baselines & \texttt{gpt-5.4-mini} & Used for the two-agent comparison between BE, prompt self-update, and RAG. \\
External stance judge & \texttt{gpt-5.4-mini} & Zero-temperature judge used for prompt baselines, the common-evaluator comparison, and the stance-expression validation. \\
AQR extraction audit & \texttt{gpt-4o-mini} & Used for the frozen extraction audit. \\
Argument strength classifier & DeBERTa-v3-large regressor & Local classifier used to score support strength before stance updating. \\
\bottomrule
\end{tabular}
\end{table}

\subsection{Prompt families and call structure}
\label{sec:exact-prompts}

Runtime values are filled from the resolved config, retrieved memory, and transcript. Table~\ref{tab:prompt-families} lists each prompt family with its inputs and structured output. The implementation stores the prompt strings used by these components. Reproducing a reported run also requires its configuration and code version.

\begin{table}[H]
\centering
\caption{Prompt families used by the LLM-facing components.}
\label{tab:prompt-families}
\scriptsize
\setlength{\tabcolsep}{4pt}
\begin{tabular}{p{0.26\linewidth}p{0.32\linewidth}>{\raggedright\arraybackslash}p{0.34\linewidth}}
\toprule
\textbf{Component} & \textbf{Inputs} & \textbf{Structured output} \\
\midrule
Argument extraction & Topic and message text & JSON list of \texttt{\{claim, evidence\_snippets\}} objects, with multi-point messages split into distinct claims. \\
Argument classification & Topic and extracted claims & JSON list of \texttt{\{polarity, support\_strength, category\_tags\}} objects in input order, with polarity fixed to affirmative or negative. \\
BE response generation & Persona, topic, current stance, retrieved memory, opponent message, and recent transcript & A concise debate response conditioned on the explicit BE stance and retrieved evidence. \\
Prompt-only baselines & Previous self-reported stance, latest opponent message, transcript, and optional retrieved context & JSON stance update plus response generation, without access to the stance update rule. \\
External stance judge & Topic, speaker name, and one generated statement & JSON stance score in $[-1,1]$, measuring expressed support for the proposition as written. \\
\bottomrule
\end{tabular}
\end{table}

\paragraph{Argument extractor prompt.} We prompt the argument extractor using the following prompt, where \texttt{\{topic\}} is the debate topic.

\begin{lstlisting}[basicstyle=\ttfamily\scriptsize,breaklines=true,columns=fullflexible,frame=single,keepspaces=true]
You are an expert conversational argument extractor.
Your task is to identify and extract distinct persuasive points from a casual chat debate regarding the topic: "{topic}".

Extraction rules:
- Extract ANY stated reason, belief, anecdote, concern, rhetorical question, or personal feeling that acts as persuasion.
- Only ignore pure filler ("hello", "how are you", "I have to go").
- If the message contains multiple distinct points, extract them separately.
- Re-phrase rhetorical questions or anecdotes into clear, standalone claims.

For each argument, identify:
1. The canonical claim: A clear, standalone sentence summarizing the point.
2. Evidence snippets: Specific quotes from the text that support this claim.

Output must be a valid JSON list of objects.
Example:
[
    {
        "claim": "Intermittent fasting helps control weight by limiting eating windows.",
        "evidence_snippets": ["I would fast one day a week and be off of on the other six days"]
    }
]
\end{lstlisting}

\subsection{Existing assets and licences}
\label{sec:asset-licences}

Table~\ref{tab:existing-assets} lists the external datasets, model families, and hosted services used in the reported experiments. We cite datasets and model families where they are introduced in the main text. The DEBATE source is \url{https://huggingface.co/datasets/seantw/DEBATE_LLM}, and the other Hugging Face entries are repository identifiers. The release does not redistribute third-party datasets, hosted-model weights, or raw proprietary API responses. Appendix~\ref{sec:code-release-plan} describes the included results and release status.

\begin{table}[H]
\centering
\caption{Existing assets used in the reported experiments, with source, licence or access terms, and redistribution handling.}
\label{tab:existing-assets}
\scriptsize
\setlength{\tabcolsep}{3pt}
\begin{tabular}{p{0.24\linewidth}p{0.29\linewidth}p{0.38\linewidth}}
\toprule
\textbf{Asset} & \textbf{Use} & \textbf{Source and licence / terms} \\
\midrule
DEBATE\_LLM dataset & Human transcript replay and pre/post opinion validation & \citet{chuang2025debate}; Hugging Face \texttt{seantw/DEBATE\_LLM}; DEBATE Dataset Research-Only License (Non-Commercial, v1.0). Not redistributed. \\
AQR / IBM argument-quality ranking data & Seed arguments and crowd-rated argument-quality labels for judgement diagnostics & \citet{Gretz2019argumentquality}; Hugging Face \texttt{ibm/argument\_quality\_ranking\_30k}; CC-BY-SA 3.0. Not redistributed. \\
DeBERTa-v3-large & Backbone for the local argument-strength regressor & \citet{deberta2023paper}; Hugging Face \texttt{microsoft/deberta-v3-large}; MIT licence. The fine-tuned scorer is released separately (Appendix~\ref{sec:code-release-plan}). \\
Qwen 3.5 9B & Open-weight generated-agent base model & Hugging Face \texttt{Qwen/Qwen3.5-9B}; Apache-2.0. Model weights not redistributed. \\
Gemma 4 E4B & Open-weight generated-agent base model & Hugging Face \texttt{google/gemma-4-E4B-it}; Apache-2.0. Model weights not redistributed. \\
OpenAI API models & Hosted generator, extraction-audit, and external-judge roles & Model identifiers in Table~\ref{tab:model-identifiers}; provider API terms apply. Model weights are not accessed or redistributed. \\
\bottomrule
\end{tabular}
\end{table}

\subsection{Code and data release}
\label{sec:code-release-plan}

The code and reproduction materials are publicly available at \url{https://github.com/joshuay1/belief-engine}, with the camera-ready version archived as release \texttt{v1.0.0}. Original code is provided under the MIT licence. The release contains the agent implementation, versioned stance-update rules, offline tests, and compact saved inputs for the reported analyses. The README gives a short setup guide, and \texttt{REPRODUCING.md} explains how to rebuild the results without model calls. In particular, \texttt{make rebuild-paper} checks the six camera-ready analysis packages, while \texttt{make rebuild-replay-stats} rebuilds the DEBATE fold statistics. A figure index maps the final manuscript figures to their saved inputs. The original reviewer snapshot remains at \url{https://anonymous.4open.science/r/belief-engine-684F}.

The fine-tuned argument-quality scorer is publicly available at \url{https://huggingface.co/joshcyang/belief-engine-argument-quality-deberta-v3-large}, under release \texttt{original-2025-12-v1}. The repository configuration pins its full revision. The scorer takes claim text alone, truncates and pads inputs to 128 tokens, and clips the raw regression output to $[0,1]$ without a sigmoid. Its model card provides download and verification instructions.

For analyses with saved model outputs, reproduction means rebuilding the reported summaries from those outputs. Live reruns additionally require access to the relevant datasets and models and may produce different responses from hosted APIs. Third-party dataset access is described in Table~\ref{tab:existing-assets}.

\section{Additional Two-Agent Topic Grids}
\label{sec:two-agent-topic-grids}

\begin{figure}[H]
    \centering
    \includegraphics[width=\linewidth]{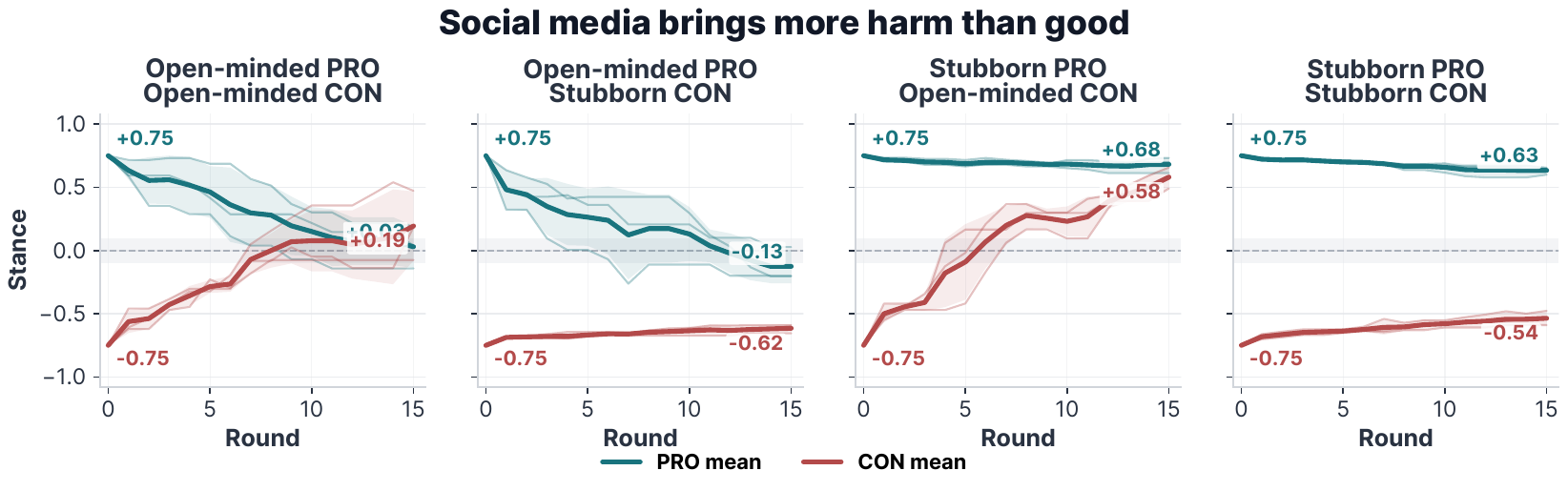}
    \vspace{2mm}
    \includegraphics[width=\linewidth]{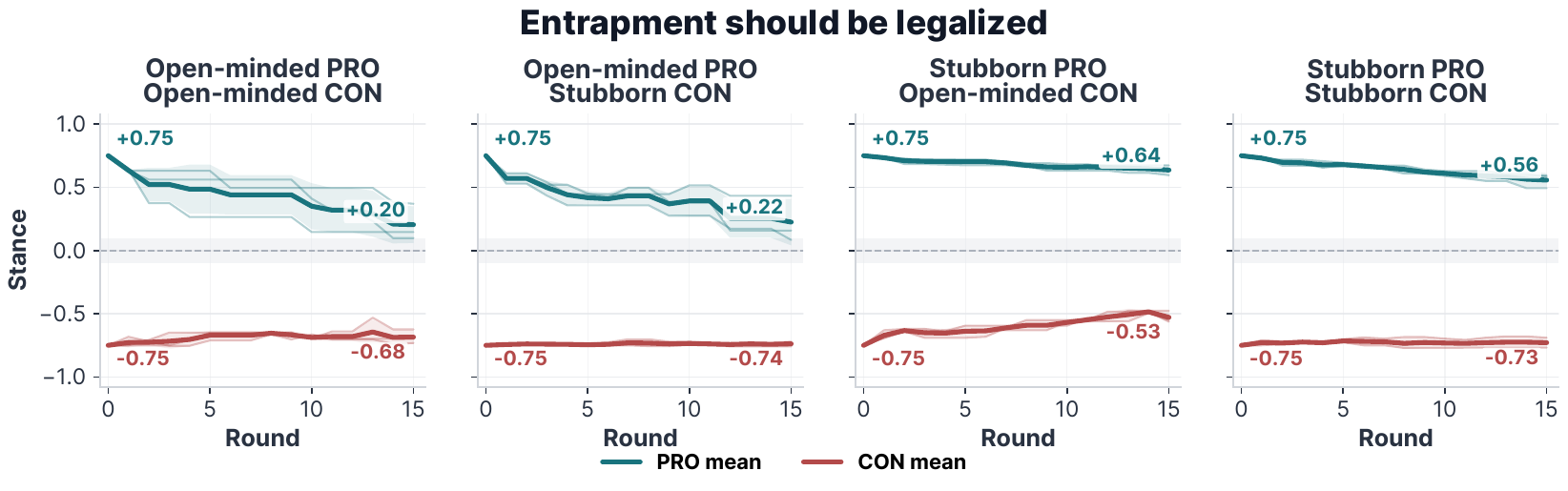}
    \caption{Additional generated two-agent topic grids. Top: Social media brings more harm than good. Bottom: Entrapment should be legalized.}
    \label{fig:two-agent-topic-grids-social-entrapment}
\end{figure}

\begin{figure}[H]
    \centering
    \includegraphics[width=\linewidth]{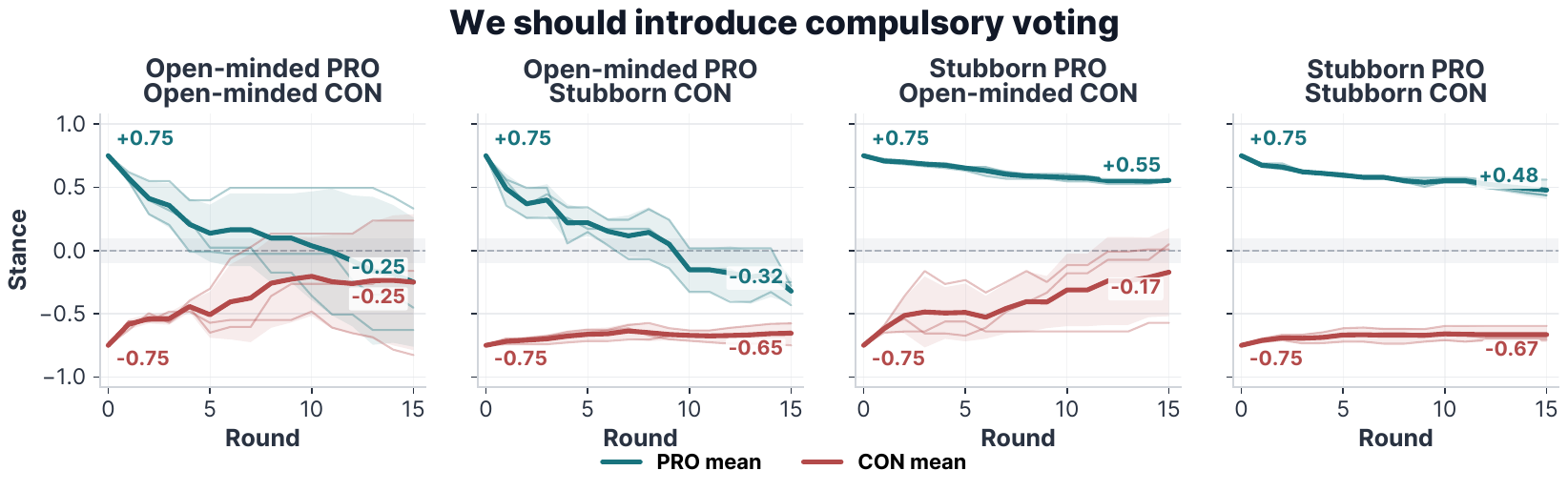}
    \vspace{2mm}
    \includegraphics[width=\linewidth]{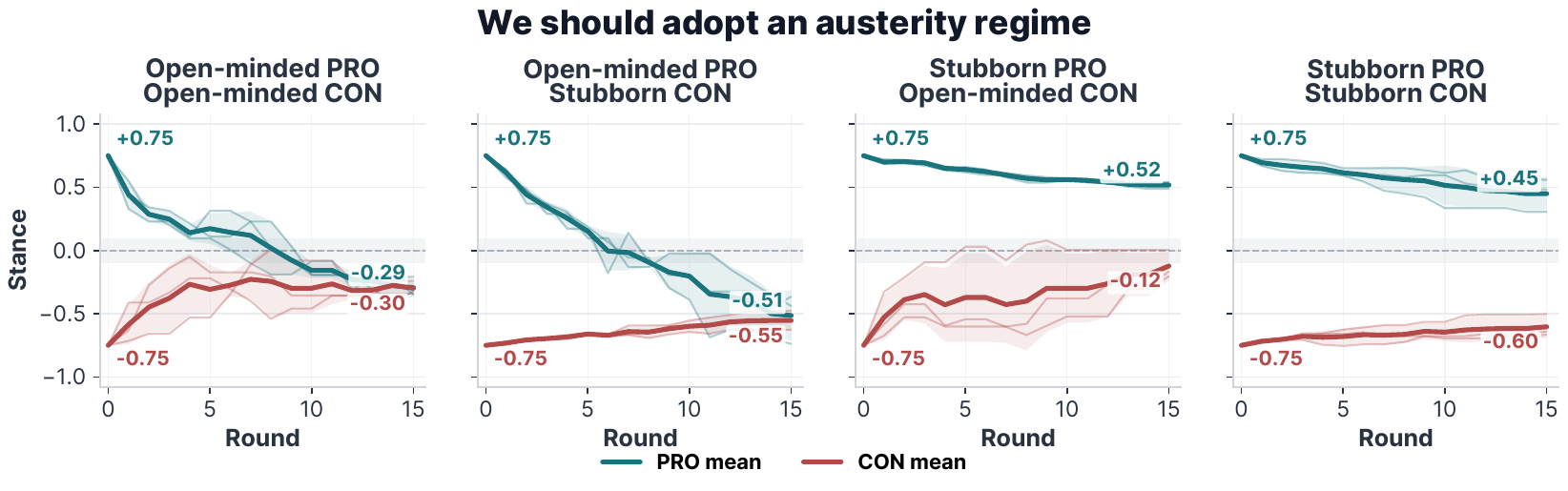}
    \caption{Additional generated two-agent topic grids. Top: compulsory voting. Bottom: austerity.}
    \label{fig:two-agent-topic-grids-compulsory-austerity}
\end{figure}

\begin{figure}[H]
    \centering
    \includegraphics[width=\linewidth]{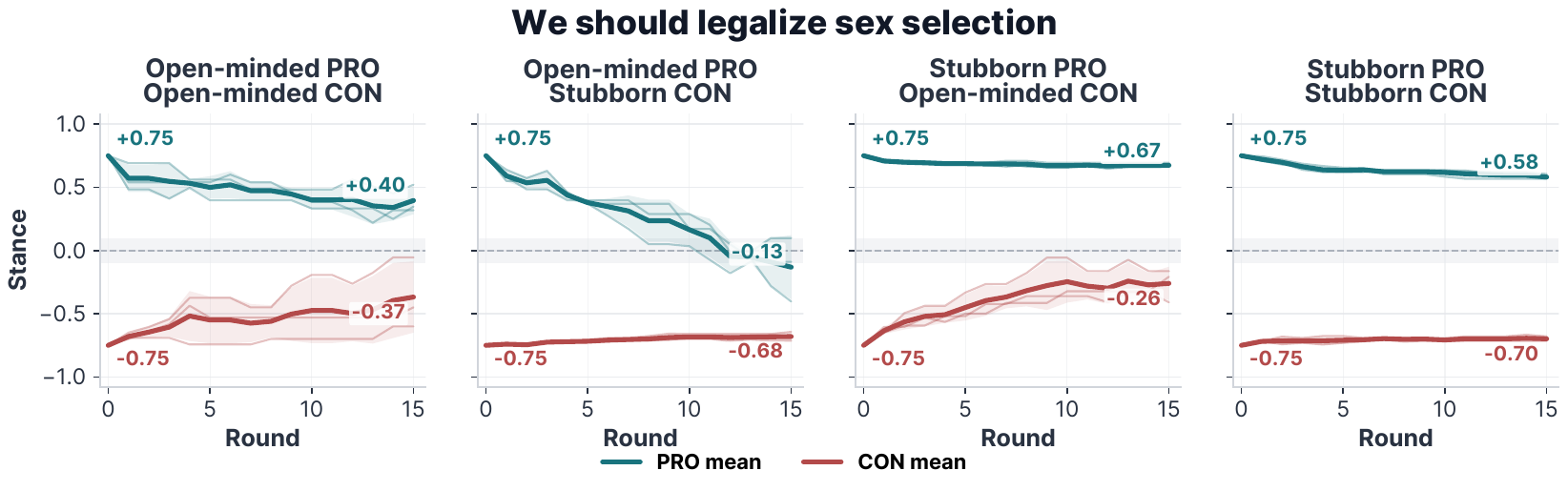}
    \vspace{2mm}
    \includegraphics[width=\linewidth]{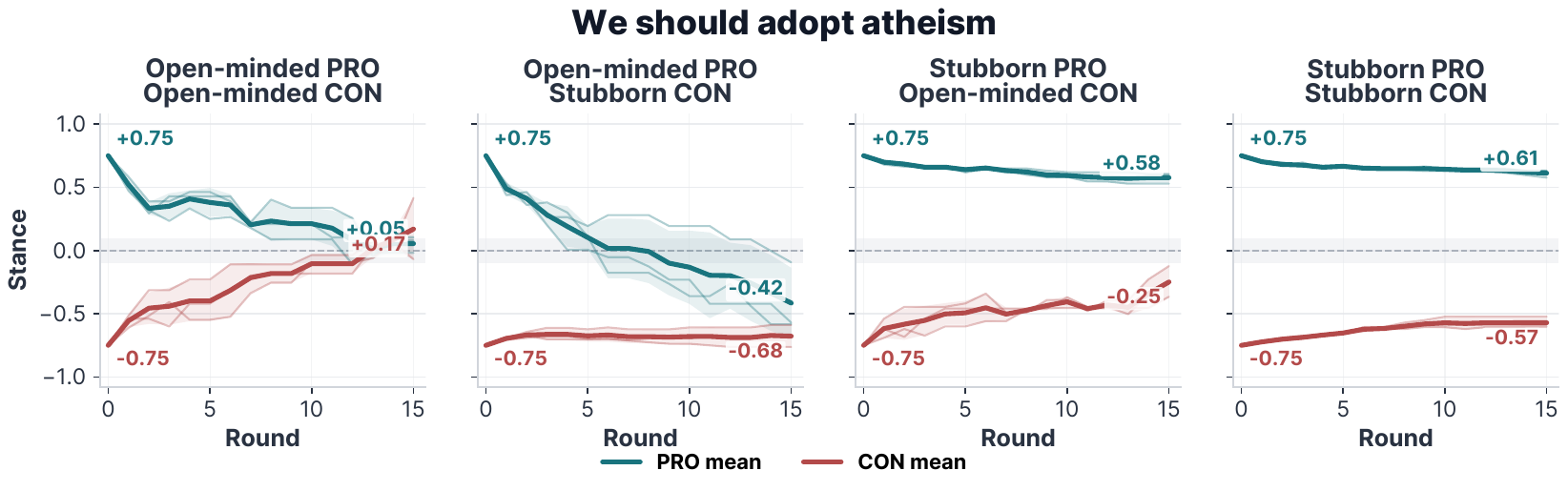}
    \caption{Additional generated two-agent topic grids. Top: We should legalize sex selection. Bottom: We should adopt atheism.}
    \label{fig:two-agent-topic-grids-sex-atheism}
\end{figure}

\begin{figure}[H]
    \centering
    \includegraphics[width=\linewidth]{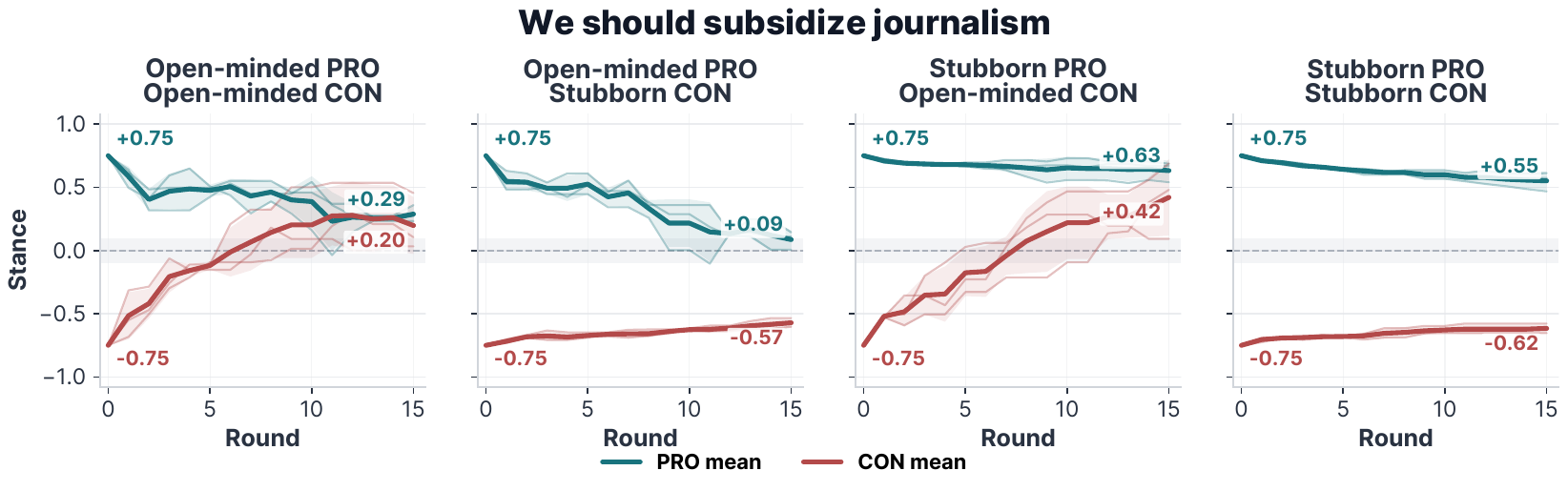}
    \vspace{2mm}
    \includegraphics[width=\linewidth]{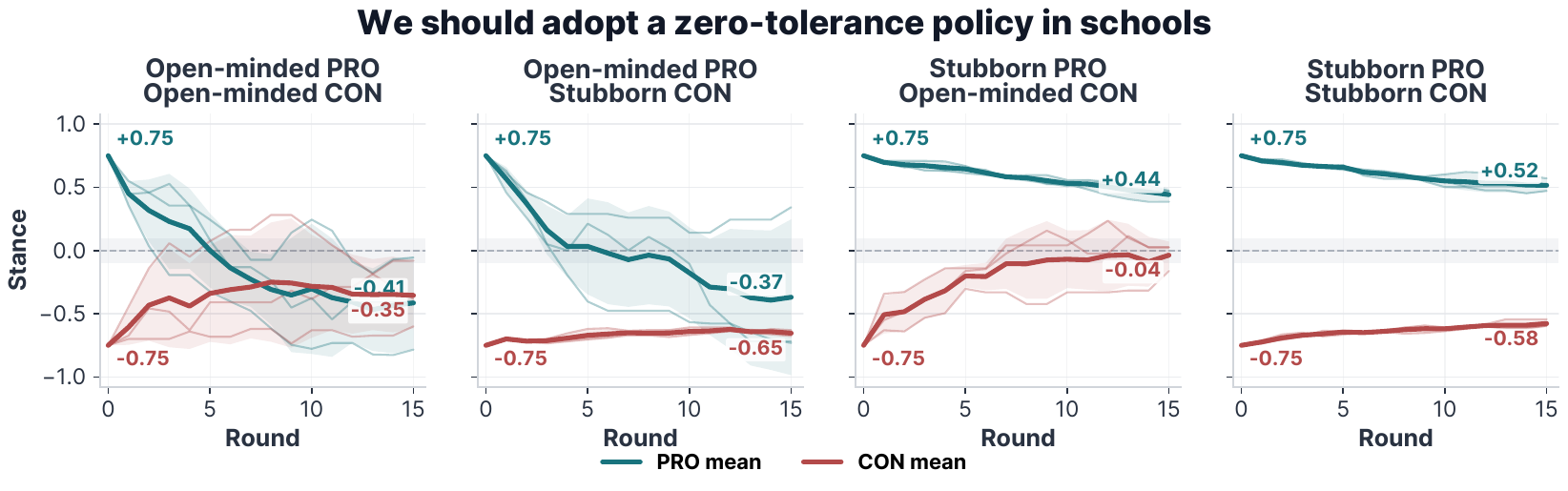}
    \caption{Additional generated two-agent topic grids. Top: We should subsidize journalism. Bottom: We should adopt a zero-tolerance policy in schools.}
    \label{fig:two-agent-topic-grids-journalism-zero-tolerance}
\end{figure}

The remaining topic-level stance grids from the generated two-agent topic run appear in Figs.~\ref{fig:two-agent-topic-grids-social-entrapment}--\ref{fig:two-agent-topic-grids-journalism-zero-tolerance}, and Table~\ref{tab:two-agent-topic-convergence} reports the corresponding topic-level convergence values.

\begin{table}[H]
\centering
\caption{Mean convergence in the ten-topic generated two-agent topic run. Convergence is the reduction in pro--con stance distance from the beginning to the end of the debate, averaged over three runs per topic--pairing cell.}
\label{tab:two-agent-topic-convergence}
\scriptsize
\setlength{\tabcolsep}{3pt}
\begin{tabular}{p{0.38\linewidth}rrrrr}
\toprule
Topic & \shortstack{Open/\\Open} & \shortstack{Open/\\Stubborn} &
\shortstack{Stubborn/\\Open} & \shortstack{Stubborn/\\Stubborn} & Mean \\
\midrule
Social media brings more harm than good & 1.255 & 1.010 & 1.399 & 0.330 & 0.999 \\
Homeopathy brings more harm than good & 0.515 & 0.279 & 0.360 & 0.099 & 0.313 \\
Entrapment should be legalized & 0.611 & 0.537 & 0.334 & 0.215 & 0.424 \\
We should adopt a zero-tolerance policy in schools & 1.091 & 1.119 & 1.021 & 0.407 & 0.910 \\
We should introduce compulsory voting & 1.305 & 1.167 & 0.774 & 0.357 & 0.901 \\
We should adopt an austerity regime & 1.436 & 1.386 & 0.860 & 0.447 & 1.032 \\
We should legalize sex selection & 0.736 & 0.951 & 0.566 & 0.221 & 0.619 \\
We should adopt atheism & 1.286 & 1.238 & 0.675 & 0.317 & 0.879 \\
We should subsidize journalism & 1.289 & 0.842 & 1.287 & 0.333 & 0.938 \\
We should adopt libertarianism & 1.313 & 1.291 & 1.336 & 0.577 & 1.129 \\
\bottomrule
\end{tabular}
\end{table}

\end{document}